\documentclass{article}

\usepackage{PRIMEarxiv}

\usepackage[utf8]{inputenc} 
\usepackage[T1]{fontenc}    
\usepackage{hyperref}       
\usepackage{url}            
\usepackage{booktabs}       
\usepackage{amsfonts}       
\usepackage{nicefrac}       
\usepackage{microtype}      
\usepackage{lipsum}
\usepackage{fancyhdr}       
\usepackage{graphicx}       
\graphicspath{{media/}}     

\usepackage{xr-hyper}
\usepackage{subcaption}
\usepackage{mathtools}
\usepackage{amsmath}
\usepackage{makecell}
\usepackage[justification=centering]{caption}
\usepackage{bm}
\usepackage{amssymb}
\title{Using Probabilistic Movement Primitives in analyzing human motion differences under Transcranial Current Stimulation
}

\author{
  Honghu Xue \\
  Institute for Robotics and Cognitive Systems \\
  University of Luebeck \\
  Luebeck\\
  \texttt{xue@rob.uni-luebeck.de} \\
  \And
  Rebecca Herzog \\
  Institute of Systems Motor Science, Department of Neurology \\
  University of Luebeck, University Medical Center Schleswig-Holstein \\
  Luebeck\\
  \texttt{rebecca.herzog@neuro.uni-luebeck.de} \\
  \AND
  Till M Berger \\
  Institute of Systems Motor Science \\
  University of Luebeck\\
  Luebeck
  \\
  \texttt{till@mariusberger.de} \\
  \And
  Tobias B{\"a}umer\\
  Department of Neurology\\
  University Medical Center Schleswig-Holstein \\
  Luebeck\\
  \texttt{tobias.baeumer@neuro.uni-luebeck.de} \\
  \And
  Anne Weissbach\\
  Institute of Systems Motor Science, Institute of Neurogenetics\\
  University of Luebeck\\
  Luebeck\\
  \texttt{anne.weissbach@neuro.uni-luebeck.de} \\
  \And
  Elmar Rueckert\\
  Chair of Cyber-Physical-Systems\\
  Montanuniversität Leoben\\
  Leoben \\
  \texttt{rueckert@unileoben.ac.at} \\
}

\begin{document}
\maketitle

\begin{abstract}
In medical tasks such as human motion analysis, computer-aided auxiliary systems have become preferred choice for human experts for its high efficiency. However, conventional approaches are typically based on user-defined features such as movement onset times, peak velocities, motion vectors or frequency domain analyses. Such approaches entail careful data post-processing or specific domain knowledge to achieve a meaningful feature extraction. Besides, they are prone to noise and the manual-defined features could hardly be re-used for other analyses. In this paper, we proposed \textit{probabilistic movement primitives} (ProMPs), a widely-used approach in robot skill learning, to model human motions. The benefit of ProMPs is that the features are directly learned from the data and ProMPs can capture important features describing the trajectory shape, which can easily be extended to other tasks. Distinct from previous research, where classification tasks are mostly investigated, we applied ProMPs together with a variant of Kullback-Leibler (KL) divergence to quantify the effect of different \textit{transcranial current stimulation} methods on human motions. We presented an initial result with $10$ participants. The results validate ProMPs as a robust and effective feature extractor for human motions. 
\end{abstract}

\keywords{Probabilistic Movement Primitives \and  Human Motion Analysis \and Transcranial Current Stimulation}

\section{Introduction}
Human motor coordination has been extensively investigated in medical research \cite{rosenbaum2009human} such as post-stroke rehabilitation \cite{hatem2016rehabilitation,gresham2004post}, Parkinson \cite{inzelberg2008visuo, plotnik2007new}, alcoholism \cite{marczinski2012effects, sullivan1995alcohol} and so on. One typical task is to examine the human motions with the presence of certain external stimuli to verify its effectiveness on human motor control. Prior to the existence of auxiliary analysis tools, examination of such effect relies on human expert via visual observation, which often results in low efficiency and less objectivity. For instance, in case of the fast movements, it would be hard for human experts to distinguish motions behavior, and high concentration easily gives rises to fatigue, consequentially incurring less accurate diagnosis. On the other hand, an auxiliary analysis tool can significantly reduce the workload, increase the efficiency as well as the accuracy. In this work, we extend the method of \textit{Probabilistic Movement Primitives} (ProMPs) \cite{paraschos2013probabilistic}, a well-established approach in robotics skill learning, to human motor coordination analysis.  

One exploratory field in medical research is \textit{noninvasive brain stimulation}, which has been long assumed as an alternative to treat neurological and psychiatric disorders. However, the brain stimulation technique is still a young research field and real applications are largely unexplored. One interesting topic is whether \textit{noninvasive brain stimulation} can influence cerebellar excitability and connectivity, consequentially leading to the changes in human motor coordination. Among all brain stimulation methods, \textit{transcranial current stimulation} (tCS) is gaining its popularity due to its easy deployablility in application \cite{filmer2014applications, datta2008transcranial, brittain2013tremor}. In the family of tCS, \textit{transcranial direct current stimulation} (tDCS) is one popular method \cite{nitsche2008transcranial, thair2017transcranial, caumo2012neurobiological}. However, tDCS shows a high variability of study protocols and applications, which registers heterogeneity of results. tDCS only exerts general effect on underlying neuron populations, without discriminating them. Therefore, other current stimulation methods serve as interesting alternatives, especially \textit{transcranial alternating current stimulation} (tACS), whose effect has been previously studied on arm movements \cite{naro2016does,naro2017effects}. Additionally, tACS can influence intrinsic oscillations, where tDCS has no influence. Besides tDCS and tACS, we also investigate the effect of \textit{transcranial random noise stimulation} (tRNS) of the cerebellum. We examine how human motions are affected under these three variants of tCS conditions.

In this paper, we applied ProMPs to evaluate the effect of tDCS, tACS and tRNS on human arm motions. Specifically, we examined whether the different stimuli approaches influenced the motion patterns as well as the duration of stimulation effect. We recorded the arm motion by inertial measurement units (IMUs) mounted on hands and wrists similar to the approach in \cite{krishna2019quantitative}. We quantified the stimulation effects by measuring the motion difference between stimulated conditions and non-stimulated conditions in eight different finger-tapping experiments, where the influence on motion coordination could be present \cite{orru2020clinical, benussi2017long}. In a finger-tapping experiment, the degree of freedom in finger tapping task is restricted, which greatly reduces the undesired effects not caused by stimulation, e.g., human exploratory behavior. The motion difference is evaluated by mapping the trajectories into feature space via ProMPs and then using a variant of \textit{Kullback-Leibler divergence} (KL-divergence) as a distance metric. In this work, we first presented an initial result with $10$ participants.

In summary, the main contributions of this paper are as follows: (i) We proposed \textit{probabilistic movement primitives} as an efficient and robust feature extractor characterizing human motions, where the features were completely learned from dataset. (ii) We quantified the effects of the different non-invasive brain stimulation approaches tACS, tDCS and tRNS using \textit{symmetric Kullback-Leibler divergence} and \textit{probabilistic movement primitives}. (iii) We presented the initial results on $10$ participants, showing our approach as an advantageous auxiliary analysis tool, and discussed the advantages and limitations. (iv) We showed a complete workflow on data collection, data post-processing, coupled with details on how \textit{Probabilistic Movement Primitives} fit on the data using IMUs.

\section{Related Work}\label{se:related work}
In this section, we review the previous approaches on analysing human motions and variants of \textit{movement primitives} applied in robot skill learning and human motion modelling. 

\subsection{Human Motion Analyses}
In order to measure the difference between the sets of trajectories, it is essential to capture some features of the trajectories, i.e., mapping the trajectory into a feature space. Several previous studies on analyzing motion difference depend either on manually-designed features of the motion trajectories \cite{markovic2020potential, kwak2020motion, bologna2016effects} or on frequency-domain analysis \cite{krishna2019quantitative, omkar2011time}. 

In \cite{bologna2016effects}, they examined the effects of cerebellar theta-burst stimulation on patients with focal dystonia by quantifying the changes in arm and neck movements. They captured the neck/arm movements by IMUs. For the head movements, they extracted the features such as angular amplitude and the maximal angular velocity, while for arm movements, the trajectory straightness, the smoothness of arm velocity curves and target overshooting were analysed. Such kinematics data were subsequently analysed by Kruskar-Wallis analysis of variance (ANOVA). In \cite{kwak2020motion}, they aimed to measure the difference of motion smoothness between a nonpathologic shoulder and a shoulder with a rotator cuff tear. They characterized such difference by measuring the angular velocity using IMUs, where they manually defined the number of peaks, peak velocity peak velocity–to–mean velocity ratio, and the number of sign reversals. However, it is noteworthy that the authors need to pre-define a set of kinematic features to characterize the motion difference, which requires good domain knowledge and can hardly be extended to other different tasks. However, ProMPs could largely mitigate this issue. We will discuss these aspects in details in Section \ref{se:disucssion}.

Another work \cite{punchihewa2020efficacy} examined the validity and reliability of IMUs in evaluating hand and trunk kinematics in a baseball-hitting scenario, the authors compared the performance between IMUs and optical motion capture systems by measuring the root mean square error across the angular displacement curves. In their paper, they also show how they derived the kinematic parameters from IMUs. The finding was that IMUs with the sampling rate of $1000$ $hz$ were sufficient in quantifying trunk and hand movement coordination in a hitting movement. Some similar findings of the efficacy of IMUs in measuring rapid movements was also verified in \cite{markovic2020potential}, where they discriminated the hand tapping motion among three different groups of females featuring different age and jobs. The difference was measured through descriptive statistical analysis on features such as motion onset time, peak acceleration/deceleration and acceleration/deceleration gradients. Based on these previous studies, we also used IMUs to capture motions from hands and wrists for its easy deployability.

There are also a variety of other work to measure the human motion difference, for instance, vision-based approaches for action segmentation or so on \cite{wang2003recent, chen2013survey}. And lots of previous work mainly focused on classification tasks, where various classifiers such as neural networks \cite{steven2018feature, hiraiwa1989emg}, linear discriminant analysis \cite{krishna2019quantitative}, support vector machines \cite{li2016lower, groh2015imu} were applied. They are normally coupled with some feature pre-processing techniques, e.g., principal component analysis. In this study, instead of classifying, we determined the degree of similarity between sets of trajectories numerically. 

\subsection{Movement Primitives}
\textit{Movement primitives} (MPs) \cite{schaal2005learning} have been extensively studied in robotics to model arbitrarily complex motor skills from both robots and humans by modularily executing multiple basic movement patterns sequentially or in parallel. 

One popular approach in the family of MPs is \textit{dynamic movement primitives} (DMPs) \cite{schaal2006dynamic, schaal2003control, ijspeert2013dynamical}. DMPs are mathematically characterized by a second-order damping system with an additional forcing term. The second-order damping system with a goal attractor asserts an asymptotic convergence to a desired pose at the end of the trajectory while the forcing term increases the model capacity to approximate trajectories of arbitrary shapes. DMPs have been widely used in robot learning, e.g., a robot pouring task \cite{tamosiunaite2011learning}. In \cite{prada2013dynamic, umlauft2014dynamic}, they applied DMPs in human-robot collaboration or interactions tasks, for instance, object hand-over. \cite{pervez2018learning} generalized DMPs from single-task learning to multiple tasks \textit{task-parameterized DMPs}, where they jointly learned the probability distribution of task-specific parameters and the shape parameters of DMPs. When inferring the trajectory for an unseen task, the phase as well as the task-specific parameters will be passed. Some other extensions for DMPs focused on online adaptation, where one of the typical tasks could be collision avoidance \cite{park2008movement, tan2011potential,hoffmann2009biologically}. 

While DMPs show great success in learning motor tasks, they only represent single elementary action. In contrast, \textit{probabilistic movement primitives} (ProMPs) \cite{paraschos2018using} model the trajectory in a probabilistic manner, which offers more flexibility than a deterministic model. Firstly, a probabilistic model can represent the motion uncertainty at every time point during the demonstrations. This uncertainty can be used to adapt control parameters \cite{paraschos2013probabilistic}. Furthermore, ProMPs can model the coupling between joints which is essential for controlling and modelling high-dimensional coupled kinematic chains such as humans and robots. A probabilistic characterization with a coupled relation also allows common probability operation such as conditioning, where a complete trajectory on all joints can be inferred from given some via-points. In the case of inferring a trajectory given via-points from different tasks, ProMPs outperforms DMPs in terms of accuracy at via-points and its adaptable variance \cite{paraschos2018using}. Similar to DMPs, ProMPs can also be extended to multiple tasks by appending task parameters to trajectory shape parameters and performing joint linear regression \cite{rueckert2015extracting}.


In robotic tasks, ProMPs can be fitted on a couple of demonstrations using imitation learning \cite{gomez2016using, maeda2014learning}. With a learned ProMP model, robots can either subsequently reproduce the demonstrated motion pattern or even improve the trajectory gradually via \textit{trajectory optimization} \cite{tamosiunaite2011learning}. \cite{paraschos2018using} verified the performance of ProMPs on benchmark tasks such as 7-link reaching task, robot hockey, playing table tennis and so on, especially in the task of table tennis, the performance of ProMPs is superior to DMPs, leading generally to smaller errors with an increased success rate. Other work also covered combining transfer learning with MP \cite{rueckert2015extracting}, where robots were trained on a few tasks and later generalize to unseen ones. \cite{ewerton2015learning, maeda2017probabilistic} further extended the idea of ProMPs to a Gaussian mixture of ProMPs for collaborative robots to coordinate the movements of a human partner. 

\subsection{Using Probabilistic Movement Primitives for Human Motion Analysis}
With the success in modelling robot motions, it is promising to extend ProMPs to human motions. Several previous work has been extensively applying MPs to model human motor skills \cite{lin2016movement}. The prior work \cite{rueckert2016probabilistic} used ProMPs to analyze the human adaptation with the presence of external perturbations by investigating the correlation between the motions of both arms and the trunk. In this work, they showed that ProMPs could sufficiently predict the complete trajectory of right arm only given trajectory of left arm the at initial phases. This finding showed ProMPs achieve a reasonable performance of in modelling the multiple trajectories in a coupled setting, which is highly-related to our work where we quantify the trajectory difference from a mixture of trajectories, i.e., from both hand and wrist. In another work \cite{kohlschuetter2016learning}, they used ProMPs on EMG data to predict knee anomalies. By learning the prior distribution of the weights on the post-processed EMG data, they passed the weights of the features including both mean and variance to the classifier. They showed a probabilistic model achieves a higher prediction accuracy than a deterministic model with no uncertainty measure. Our work is related to them in a way that we measured the trajectory difference in a probabilistic manner (with uncertainty measure), potentially being more accurate than a deterministic model.

Some other work also proposed using MPs for motion motion. \cite{lim2005movement} proposed using MPs to generate natural, human-like motions with a framework combining dynamic models and optimization. For instance, \cite{Rueckert2013b} suggested a movement primitive representation as a generalized case of DMPs to implement shared knowledge in form of learned synergies, where the learned synergies summarized the muscle excitation patterns and enabled transferring to other tasks given muscle signal in musculoskeletal systems.

Based on the previous studies, we proposed using ProMPs for modelling the coupled movements from hand and wrist to determine the effect of \textit{transcranial current stimulation}. To the best of our knowledge, this work is the first attempt to apply ProMPs to investigate effect on non-invasive brain stimulation approaches.

\section{Methods}\label{se:Methods}

In this section, we provide a mathematical formulation on \textit{Probabilistic Movement Primitives} (ProMPs) and subsequently show how to fit the motion trajectories using ProMPs. We start by introducing the definition of \textit{time-series data}, on which ProMPs will be fitted. Afterwards, we extend a single time-step case to multi time-step case, i.e., a set of complete trajectories. Finally, we generalize a complete trajectory to coupled trajectories.

\subsection{ProMPs as a Probabilistic Time Series Model}\label{se:Time_series_data}

A \textit{trajectory} $\tau$ is a sequence of observations $y$. Mathematically, we define a complete trajectory as $\bm{\tau_{j}} = [y_{0,j},...,y_{T,j} ]$, where $y_{t,j}$ denotes the observed measurements at time point $t$ and the trajectory $j$ has the length of $T$. We assume scalar observations where $y_{t,j} \in \mathbb{R}^{1}$ and later we will extend to multi-dimensional cases. Now, we consider a set of trajectories with the same length $T$, and arrive at the data form $ \bm{S}= \{ y_{0,1}, y_{1,1}, ...,y_{T,n} \} $, where $n$ refers to the number of demonstrations or trajectories. The goal is to derive a probabilistic time series model from these $n$ trajectories denoted by $p(\bm{\tau})$.

\subsection{Probabilistic Time Series Model on single time step} \label{se:single_time_step_model}
We first start with a probabilistic characterization on $\bm{S}$ the for a single time step using Gaussian distribution. For simplicity, we use the notation $\bm{y_{t}}$ to represent a set of $n$ observations at time step $t$, i.e., $\bm{y_{t}} = [[y_{t,1},y_{t,2},...,y_{t,n}]] \in \mathbb{R}^{1 \times n}$. We assume a Gaussian distribution on the observations with the mean as a linear combination of non-linear features $\phi$ and a fixed variance, shown as below:
\begin{equation}
\label{eq:single_time_step}
P(\bm{y_{t}}|\bm{w}) = \mathcal{N}(\bm{y_t}|\bm{\phi_{t}} \bm{w}, \sigma_{y}^{2}),
\end{equation}
where $\bm{\phi_{t}} \in \mathbb{R}^{1 \times M}$ denotes the feature vector at the time step $t$ and $\bm{w} \in \mathbb{R}^{M \times n}$ denotes the weight vector to be learned. In this case, we assume the number of used basis functions to be $M$. The standard deviation of the observations $\sigma_{y}$ can be interpreted as a noise term.

The common choice of the feature $\phi_{t}^{i} \in R^{1} \;, \forall i \in [1,...,M]$ varies for stroke-based movements or rhythmic movements \cite{paraschos2013probabilistic}. In our case, the finger-tapping trajectory are firstly segmented to a set of point-to-point trajectories, where the stroke-based movements are appropriate choices, shown as follows: 
\begin{equation}
\label{eq:bf}
\begin{aligned}
\phi_{t}^{i} &= \exp{\left(-\frac{(z_t - c_i)^2}{2h_i} \right)}, \\
\bm{\phi_{t}} &= \frac{1}{\sum_{i=1}^{M}\phi_{t}^{i}}[[\phi_{t}^{1},...,\phi_{t}^{M}]],
\end{aligned}
\end{equation}
where $c_i$ refers to the center and $h_i$ refers to the bandwidth of the $i$-th basis function, $z_t= \frac{t}{T} \in [0,1]$ is the \textit{movement phase}, which is a generalization of time that allows for generating motions of arbitrary duration. 
Until now, we describe how to map measurements of single time step $\bm{y_{t}}$ into a feature space $\phi$. Then, we show how to compute the optimal ProMPs model parameters $\bm{w^{\star}} \in \mathbb{R}^{M \times n}$ on that single time step data $\bm{y_{t}}$. In principle, $\bm{w^{\star}}$ can be learned via \textit{expectation maximization} \cite{rueckert2015extracting} or simply through \textit{damped least-square regression}, i.e., \textit{ridge regression}.

\begin{equation}
\label{eq:pseudo_inv_single_step}
\bm{w^{\star}} = (\bm{\phi_{t}^{T}} \bm{\phi_{t}} + \lambda \bm{I})^{-1} \bm{\phi_{t}^{T}} \bm{y_{t}},
\end{equation}

where $\bm{I}$ is the identity matrix and the additional term of $\lambda$ avoids the singularities which can be caused by small or poorly-sampled datasets. The poorly-sampled data refers to the importance of motion variance in the data (copies of the same demonstrations would result in singular matrices). 

\begin{figure}[t]
\centering
\includegraphics[width=0.7\textwidth]{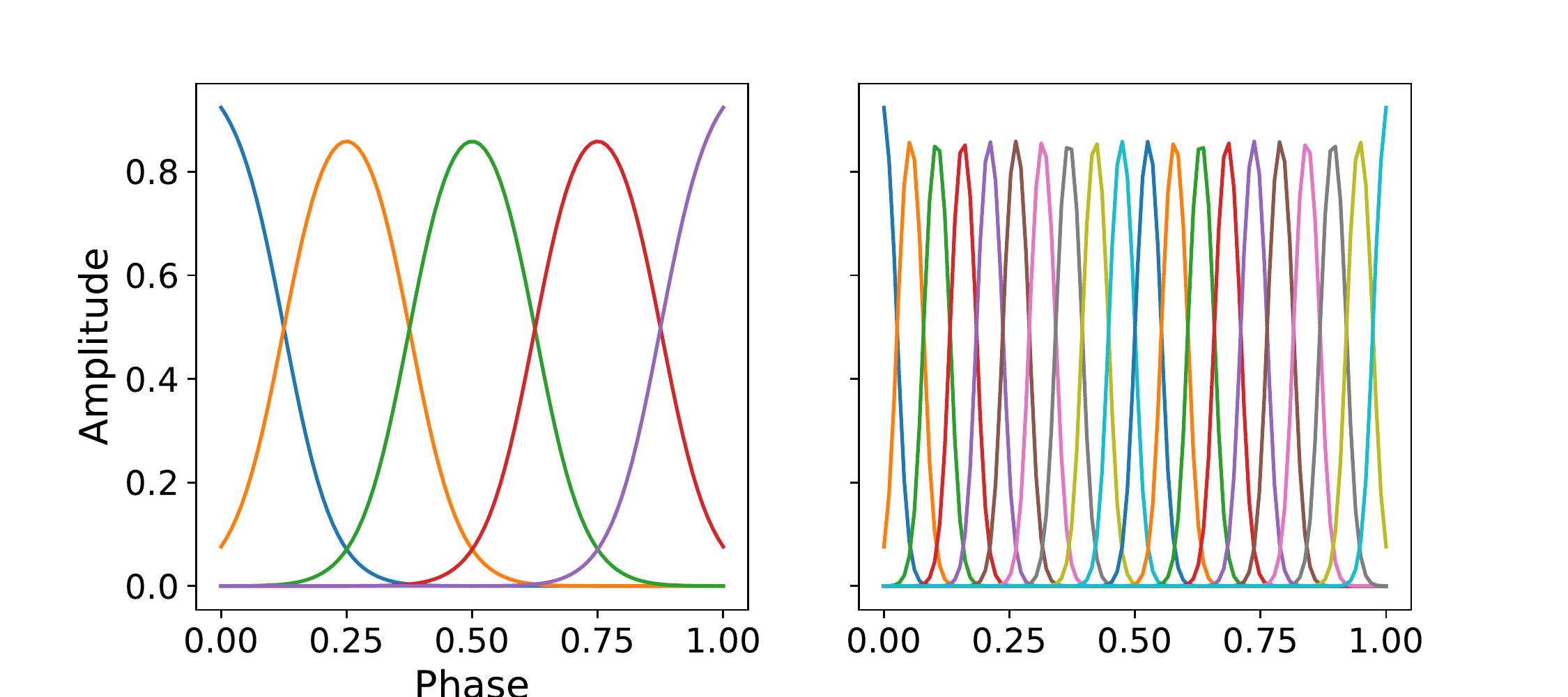}
\caption{Illustration of the distribution of basis function. Left: $M=5$, Right: $M=20$. The bandwidth parameter is computed by $h = 0.2 \cdot (c_{i+1}-c_{i})^{2}$ to  ensure an overlap between neighboring basis functions.}
\label{fig:bf}
\end{figure}


\subsection{Probabilistic Time Series Model on multi-time-step data and multi-dimensional observations} \label{se:multi_time_step_model}
The previous subsection describes how to learn the weights from multiple demonstrations on a single time step. In this part, we extend it to multi-time step setting, i.e., learning the weights from multiple complete trajectories. We assume the observations in each time step are \textit{i.i.d.}.
\begin{equation}
\label{eq:multi_time_step_independence}
\begin{aligned}
    p(\bm{\tau | \bm{w}}) &= p(\bm{y_{0}}, \bm{y_{1}},..., \bm{y_{T}} | \bm{w}) \\
    &= \prod_{t=0}^{T} p(\bm{y_{t}} | \bm{w})\\
    &= \prod_{t=0}^{T} \mathcal{N}(\bm{y_{t}}| \bm{\phi_t} \bm{w}, \sigma_{y}^2) \\
    & = \mathcal{N}(\bm{\tau}| \bm{\Phi} \bm{w}, \sigma_{y}^2 \bm{I}),
\end{aligned}
\end{equation}
where the matrix $\bm{\Phi} = [\bm{\phi_0}, \bm{\phi_1}, ...,\bm{\phi_T}] \in \mathbb{R}^{T \times M}$. The optimal weights are computed in the same way as in Equation \ref{eq:pseudo_inv_single_step} by replacing $\bm{\phi_t}$ with $\bm{\Phi}$.

In the above parts, we introduce how to fit ProMPs on a set of trajectories in the case of scalar observations, where $y_{t,j} \in \mathbb{R}^{1}$. Here, we extend the setting to multi-dimensional observations, i.e., $\bm{y_{t,j}} \in \mathbb{R}^{D}$ and $\bm{\tau} \in \mathbb{R}^{(T \cdot D) \times n}$, where $D$ is the number of dimensions. In our case, we have a $6$-dimensional observation for each time-step, respectively $x$, $y$ and $z$-axis of both hands and wrists, then the corresponding feature vector $\bm{A}$ is defined as follows:
\begin{equation}
\label{eq:coupled_phi}
\bm{A} = 
\begin{pmatrix}
    \bm{\Phi^{[hand,x]}} & 0 & \dots & 0 \\
    0 & \bm{\Phi^{[hand,y]}} & \dots & 0 \\
    \vdots & \vdots & \ddots & \vdots \\
    0 & 0 & \dots & \bm{\Phi^{[wrist,z]}}
\end{pmatrix} \in \mathbb{R}^{(T \cdot D)  \times (M \cdot D)}.
\end{equation}
With a multi-dimensional formulation, it is possible to learn the correlation between each dimension, which enables inferring the measurement of one dimension given the observations from other dimensions. The optimal weights $\bm{W^{\star}} \in \mathbb{R}^{(M \cdot D) \times n}$ can be computed as follows:
\begin{equation}
\label{eq:pseudo_inv}
\bm{W^{\star}} = (\bm{A}^{T} \bm{A} + \lambda \bm{I})^{-1} \bm{A}^{T} \bm{\tau}.
\end{equation}
\subsection{Modelling a distribution over a set of trajectories}\label{se:trajectory_prediction}
Given a set of trajectories $\bm{\tau}$, we can compute the parameter posterior $p(\bm{W})$ with the knowledge of $W^{\star}$ of each single trajectory. We assume $p(\bm{W})=\mathcal{N}(\bm{W}|\bm{\mu_{w}}, \bm{\Sigma_{w}})$, where $\bm{\mu_{w}}$ and $\bm{\Sigma_{w}}$ are computed by collapsing the dimension of $n$.
A distribution over a set of trajectories $p(\bm{\tau})$ can be computed via the marginal over the joint distribution of $p(\bm{\tau}, \bm{W})$, i.e.,
\begin{equation}
\label{eq:P(tau)}
\begin{aligned}
p(\bm{\tau}) &= \int p(\bm{\tau}, \bm{W}) d\bm{W} \\ 
&= \int p(\bm{\tau}|\bm{W}) p(\bm{W}) d\bm{W} \\
&= \int \mathcal{N}(\bm{\tau}|\bm{\Phi} \bm{W}, \bm{\Sigma_{y}}) \mathcal{N}(\bm{W}|\bm{\mu_{w}}, \bm{\Sigma_{w}}) d\bm{W} \\
&= \mathcal{N}(\bm{\tau}|\bm{\Phi} \bm{\mu_{w}}, \bm{\Phi} \bm{\Sigma_{w}} \bm{\Phi^{T}} + \bm{\Sigma_{y}}).
\end{aligned}
\end{equation}
With that, we show a probabilistic reconstruction of the demonstrated trajectory set.

\subsection{Measures for computing motion similarity}\label{se:distance_metric}
One goal of this paper is to quantify the effect of three \textit{non-invasive brain stimulation} approaches tDCS, tACS and tRNS on the human motions. For that, metrics are required to describe the distance between two sets of trajectories, i.e., $p(\bm{\tau}^{[A]})$ and $p(\bm{\tau}^{[B]})$. According to \cite{stark2017comparison}, \textit{Kullback–Leibler divergence} is the best option for point to point motions, shown as below: 

\begin{equation}
\label{eq:KLD}
\begin{aligned}
D_{KL}(P\|Q) &= \sum_{x \in \mathcal{X}} P(x)\log\left ( \frac{P(x)}{Q(x)} \right ), \\
\end{aligned}
\end{equation}

where $P(x)$ and $Q(x)$ are two probability distributions defined on the same probability space $\mathcal{X}$. In the case where $P(x)$ and $Q(x)$ are equivalent, $D_{KL}(P\|Q) = 0$, and KL-divergence increases monotonically with the discrepancy between two probability distributions, namely $D_{KL} \in \left [0, +\infty \right )$. Note the KL-divergence is asymmetric as $D_{KL}(P\|Q) \neq D_{KL}(Q\|P)$, which is an undesired property for a distance metric. Therefore, to enforce symmetricity, we simply used a trick as in \cite{symKL}, defined as symmetric Kullback-Leibler divergence $D_{KLS}$ shown below:

\begin{equation}
\label{eq:KLD_sym}
\begin{aligned}
D_{KLS}(P\|Q) & = D_{KLS}(Q\|P) \\
        & =\frac{1}{2} (D_{KL}(P\|Q) + D_{KL}(Q\|P)).
\end{aligned}
\end{equation}

\begin{figure}[tb]
\centering
\includegraphics[width=0.8\textwidth]{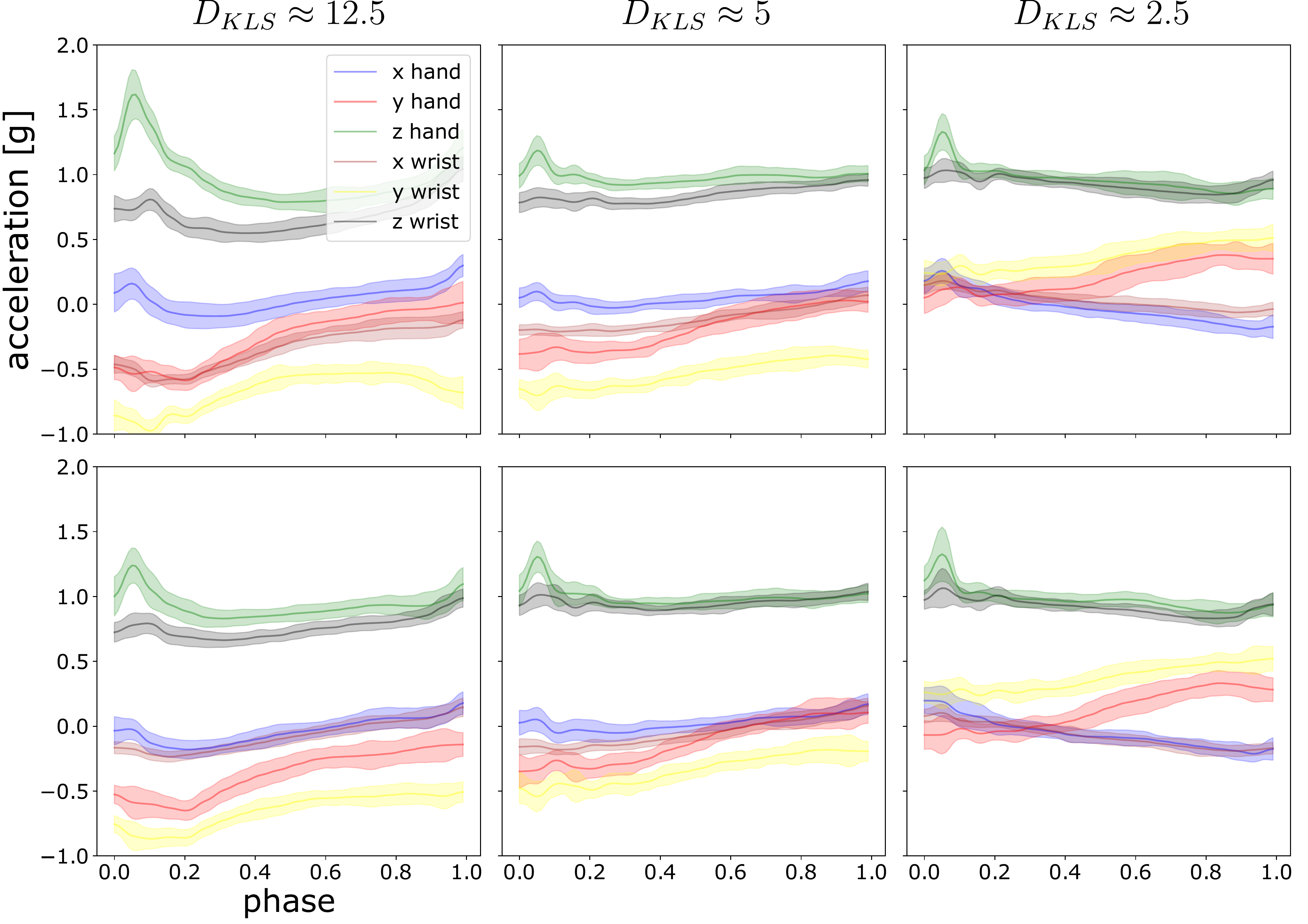}
\caption{This illustrations shows three examples of the divergent value of $D_{KLS}$ and the degree of difference in trajectory space. The post-processed acceleration profiles are fitted by ProMPs, with the shaded area being $1\sigma$ confidence level. In each panel, six motion trajectories of hand and wrist in \textit{x}, \textit{y} and \textit{z}-axis are shown. The top rows are compared with the lower rows to compute $D_{KLS}$. From left to right, the distance measure $D_{KLS}$ between the two sets is decreasing.}
\label{fig:KL_value_correspondence}
\end{figure}

In our case, we recorded $x$, $y$ and $z$-axis for both hand and wrist, altogether $6$ trajectories for one experiment. The symmetric KL-divergence was computed as the averaged $D_{KLS}$ over $6$ axes. Figure \ref{fig:KL_value_correspondence} presents a sketch of the correspondence of numerical KL divergence value to its probability distribution discrepancy. In the case of (coupled) trajectories, we defined the divergent value of two sets of trajectories $\bm{\tau_{1}}$ and $\bm{\tau_{2}}$ as the mean value by averaging the number of the discrete time points in \textit{phase}, shown as follows:

\begin{equation}
\label{eq:KLD_sym_traj}
D_{KLS}(\bm{\tau_{1}}\|\bm{\tau_{2}})  = \frac{1}{T} \sum_{t=0}^{T} D_{KLS}\left(\bm{y_{t}^{1}} \| \bm{y_{t}^{2}} \right),
\end{equation}
where $\bm{y_{t}^{1}}, \bm{y_{t}^{2}} \in \mathbb{R}^{6 \times n} $ refer to two sets of 6-dimensional measurements at the discrete time point $t$ in our case, with each each set containing $n$ measurements.

\section{Experiment Design}\label{se:experiment_design}
In this section, we present the detailed description on experimental design to measure the effect of tACS, tDCS and tRNS on human motions. In addition, we introduce the complete workflow on data post processing.
\subsection{Sensors}\label{se:sensors}
\begin{figure}[tb]
\centering
\includegraphics[width=0.6\textwidth]{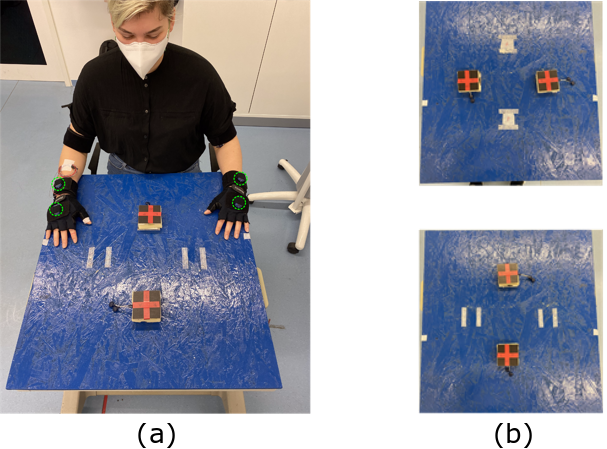}
\caption{(a) Illustration of sensor montage and experiment environments, altogether four IMUs are mounted on subject's wrists and hands on both sides, marked in green. (b) Location of touching pads for different movement direction. Left: left-right tapping direction, Right: posterior-anterior tapping direction. The touch pads provide binary signals which are used to segment the motion signals into individual patters for training the ProMPs, see Section \ref{se:data_process}.}
\label{fig:sensors}
\end{figure}

We captured all participants' motion trajectories by IMUs, i.e., Myon aktos-t sensors. Each aktos-t transmitter includes 3-axial sensors for accelerometer, gyroscope and magnetometer. The transmitters are combined with the aktos EMG system for up to $32$ channels. We set the sampling frequency of accelerometer, gyroscope and magnetometer to be $2000\;hz$. We also placed two accelerometers as pressure sensors under the touching pad to split the $1$-minute recording into movement segments from one pad to the other and vice versa. 
We mounted totally four sensors respectively on both hands and wrists to capture the arm motion. The recorded sensor profiles will be post-processed for motion difference analysis. The detailed configuration on how sensors are placed is shown in Figure \ref{fig:sensors}(a). 

\subsection{Experimental Tasks}\label{se:exp_tasks}
In our design, we chose a finger-tapping motion for examining the motion difference under different tCS approaches. In the finger-tapping motion, the subject is asked to perform tapping between two pads with fixed location repetitively. \cite{orru2020clinical, benussi2017long} also design similar tasks for investigating the effect of tCS variants on coordination. The benefit of a finger-tapping experiment lies in its simplicity for repetitive demonstrations, which is easy for participants to repeat multiple times. Moreover, the degree of freedom of the finger-tapping motions is low and therefore the motion difference can be mostly counted as the effects of tCS approaches. Thirdly, the repetitive motion is also beneficial to ProMPs from a probabilistic perspective, it is more accurate to characterize a probability distribution given more data samples (demonstrations). In total, we defined eight different tapping patterns, (we also call each pattern as unit experiment for the rest of the paper) for each stimuli approach so that the effect of stimuli can be observed in a comprehensive manner. Each experimental setting is shown in Table \ref{tab:8_exp}. Among them, we distinguished different conditions of tapping with favorable hand, different tapping directions and tapping speed. Below we will discuss the settings for these three categories in detail.

\textbf{Motion Direction:} We set two motion directions, respectively tapping in left-right direction and in anterior-posterior direction, to get the possibility to differentiate between side-dependent and side-independent motions. Figure \ref{fig:sensors}(b) shows the detailed setting, where the tapping pads are $30\;cm$ away from each other. 

\textbf{Motion Speed:} We distinguished two movement patterns, i.e., rhythmic and rapid movements patterns. For rhythmic movements, the subject first followed the beep signal corresponding to $1.5\; hz$ to tap for $15$ seconds. The beep signal then vanished and the subject was instructed to go on with the same rhythm for additional $30$ tapping cycles. The total recording time including the beep signal endures one minute. In the rapid motion case, the subject was instructed to perform tapping as fast as possible for $30$ tapping cycles, which lasted from $15$ to $30$ seconds. 

\textbf{Motion with favorable hands:} We also asked all the subjects to perform the experiments with both hands. Thus, we could also examine to what extent the stimulation approach could impact on favorable/non-favorable hands.

\subsection{Experimental Protocol}\label{se:exp_protocol}
Each participant got measured on four days in total. Each of the four days was set with one week apart from the others to avoid carry-over effects. On each day, the participant received only one type of stimulation, i.e., tACS, tRNS or tDCS, which took up three days. We also included a Sham stimulation, which only mimicked the sensory sensation of real tDCS but did not cause any plasticity effects, as it included only $30$ sec. of real stimulation. The stimulation approach were kept unknown from the participant to allow a fair analysis. The recordings of Sham allows us to distinguish whether the difference in motion pattern arises from the neuronal plasticity, i.e., the long lasting changes in the signal transmission induced by the electric stimulation, or repetition only. The order of the tCS protocols including Sham was randomized.

In order to check the influence of the duration of each stimuli, we first familiarized the participant with the eight experiments before the stimuli was activated, 
All participants could initially practice all eight experiments for a first trial without any stimulus being active. This phase is denoted by the term \textit{Erst}. Going through the experiments once helps reduce the variance of motion patterns for the coming experiments by warming up the participants, as the experiment subject can show different motion patterns, not caused by stimuli but due to the unaccustomedness to the experiment. 

Around $45$ minutes after \textit{Erst}, we assume the participants were already familiar with the on-going experiments and also reach a stable state for further experiments, the participants woulf perform all the experiments again (without stimuli), we name the second run \textit{Prae}, which serves as the non-stimulated baseline. After \textit{Prae}, the participant accepted the stimuli and we periodically recorded the motion data around every $35$ minutes until the maximal duration of $115$ minutes are reached. Altogether, we have three recordings after stimuli, namely $post \, 1$, $post \, 2$ and $post \, 3$. Table \ref{tab:exp_schedule} in Supplementary Material elaborates the procedure.


\subsection{Participant Information}
In this current work, we included an initial study of $10$ healthy subjects, all are right-handed. Among them, $8$ subjects are females. The mean age is $23$ years with a range of $20$-$31$ years. All subjects are without reported somatic or psychiatric diseases. The order of each participant's experiments are shuffled and shown in Table \ref{tab:exp_schedule2} in Supplementary Material. The overall number of experiments is $10 \cdot 4 \cdot  5 \cdot 8 = 1600$.


\subsection{Data Post-processing}\label{se:data_process}
\begin{figure}[bt]
\centering
\includegraphics[width=\textwidth]{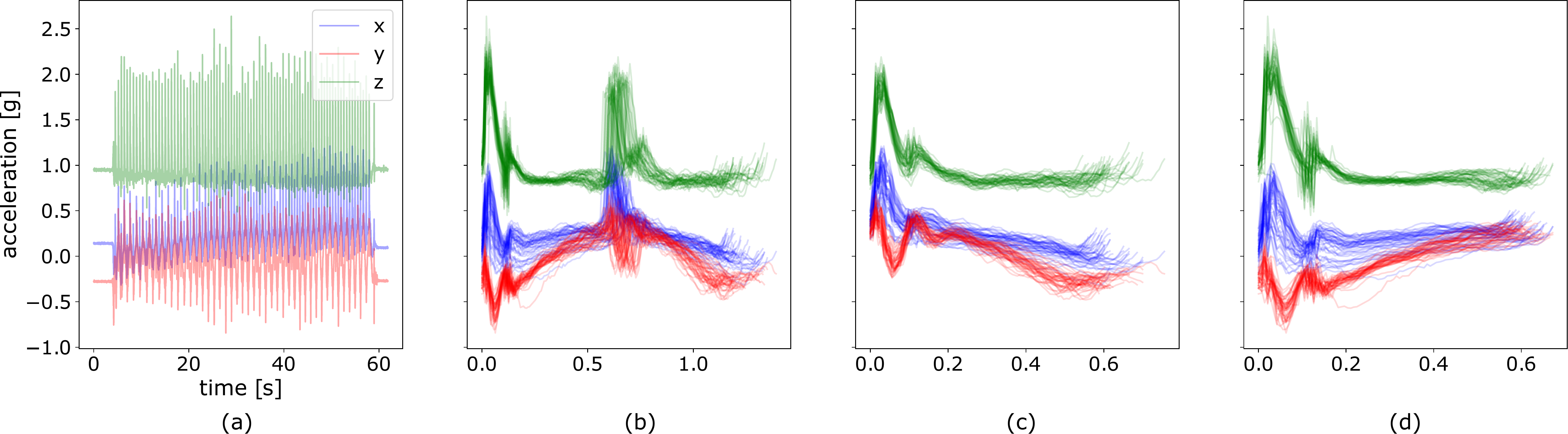}
\caption{An illustration of raw acceleration profile and segmented data. (a) Raw hand acceleration profile of a random subject in one experiment. (b) Segmented data for a complete movement from the touching pad $1$ to the pad $2$ and again to pad $1$, concatenating both inward movements and outward movements. (c) Segmented motion from the pad $1$ to the pad $2$, namely, inward movements. (d) Segmented motion from the pad $2$ to the pad $1$, outward movements.}
\label{fig:segmented_data}
\end{figure}

Before applying ProMPs to analyze the motion difference, we first post-processed the data. This includes (i) data segmentation (ii) alignment and normalization of data over time, i.e., mapping time to the \textit{movement phase} as mentioned in Section \ref{se:single_time_step_model}.
In data segmentation, we segmented the complete raw trajectory into two sets of point-to-point trajectories, namely inward movements and outward movements respectively. Inward movements refer to the direction where the angle between upper arm and forearm decreases with the motion, vice versa for outward movements. As an illustration, an inward movement for left arm is from left pad to right pad and upper pad to lower pad in Figure \ref{fig:sensors}(b) assuming the subject's location is to the lower part of the table. 
By doing this, we decomposed the motion into stroke-based motions. In these stroke-based motions, we could fit our ProMP model and further compare the difference of non-stimulated patterns against stimulated ones. The segmentation was performed given the statistics of the accelerometer under the touching pad and is shown in Figure \ref{fig:segmented_data}. 

\begin{figure}[b]
\centering
\includegraphics[width=0.6\textwidth]{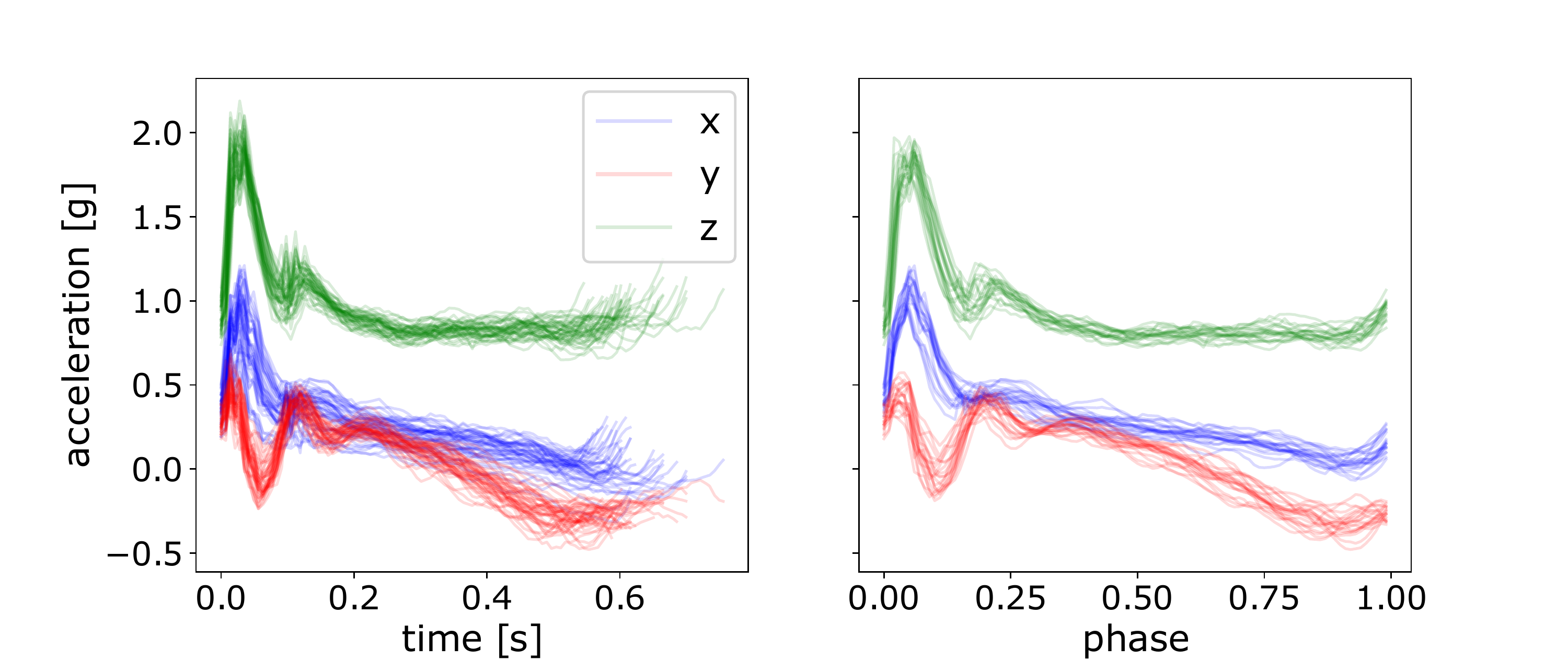}
\caption{Perform time normalization over all the segmented acceleration data with various time length. Left: only segmented data, but not normalized, Right: segmented and normalized data of the last $20$ strokes.}
\label{fig:time_aligned_data}
\end{figure}

After segmentation, we could see each segment with various time duration. Since $\bm{\Phi}$ is the pre-computed matrix used in Equation \ref{eq:multi_time_step_independence}, the \textit{phase} of each segmented trajectory must be of the same resolution. And one simple way to achieve that is to perform time alignment and normalization on each segmented piece and then convert the time $t$ to the phase $z_t$ as in Equation \eqref{eq:bf}. It is shown in Figure \ref{fig:time_aligned_data}. For fitting model parameters of ProMPs, we took $20$ segments from the last twenty-first stroke to the last second stroke. The last stroke was excluded for fitting to avoid subjects' unintended movement when hearing the stop signal. By this, we have totally $10 \cdot 5 \cdot 8 \cdot 20=8000$ strokes of movements for each stimulation approach.

\subsection{Measuring the Effect of Stimulation Methods}\label{se:kl_compute}
In this subsection, we introduce how we characterized the finger-tapping motions using ProMPs and the motion difference measure $D_{KLS}$ as defined in Section \ref{se:Methods} to measure to what extent and duration of each stimulation approach on motion patterns. We first fit the ProMPs on the segmented and time-aligned data for each of $1600$ experiments, i.e., $1600$ sets of trajectories, as described in Section \ref{se:exp_tasks}. The fitted result showed a probability distribution of the model parameters $\bm{W}$ via $\bm{\mu_{w}}$ and $\bm{\Sigma_{w}}$. We then mapped the feature $\bm{W}$ back to the trajectory space as shown in Section \ref{se:trajectory_prediction} so that a probabilistic characterization of the post-processed trajectories $\bm{\tau}$ was available. We denote such reconstructed trajectories from ProMPs as $\bm{\tau^{\prime}}$. To measure the duration of each stimulation's effect, we computed the mean \textit{symmetric KL-divergence} of \textit{prae} against \textit{post 1}, \textit{post 2} ,\textit{post 3} over all experimental subjects and eight experimental settings in the space of $\bm{\tau^{\prime}}$, namely: 

\begin{equation}
\label{eq:KLD_for_effects}
\begin{aligned}
&D_{KLS}(\bm{\tau^{\prime}_{[o=prae],p,q}}||\bm{\tau^{\prime}_{[o=post\;1],p,q}}),\\
&D_{KLS}(\bm{\tau^{\prime}_{[o=prae],p,q}}||\bm{\tau^{\prime}_{[o=post\;2],p,q}}),\\
&D_{KLS}(\bm{\tau^{\prime}_{[o=prae],p,q}}||\bm{\tau^{\prime}_{[o=post\;3],p,q}}),
\end{aligned}
\end{equation}

for each $p$ and $q$ on all available data for inward movements and outward movements respectively, where $o$ refers to the four experimental phases of \textit{prae}, \textit{post 1}, \textit{post 2} and \textit{post 3}, and $p$ denotes the index of different unit experiments. The index $q$ stands for the number of participants. We excluded \textit{Erst} for statistical analysis as the purpose of \textit{Erst} is to familiarize the subject with experiment and the subjects are likely to perform exploratory behaviors for each unit experiment.

\section{Results}\label{se:experiments}
In this section, we explain in detail how we computed the trajectory difference using \textit{probabilistic movement primitives} (ProMPs) to reveal the effect of different \textit{transcranial current stimulation} methods on human arm motion. We present comprehensive results using all of the available data to show that ProMPs togerher with the \textit{symmetric KL-divergence} can be used for characterizing trajectory differences in brain stimulation studies. We proceed in a progressive manner by answering the following questions: 
(i) Which sensor profile is best suited to analyze stimulation methods? (acceleration, velocity or displacement profile) 
(ii) What is the reconstruction error using ProMPs?
(iii) Can ProMPs be used to detect outliers? 
(iv) How are the effects of tRNS, tDCS and tACS over the finger-tapping motion?
(v) Can time-specific differences be detected on a millisecond time scale?

\subsection{Which sensor profile is best suited to analyze stimulation methods?}\label{se:fit_target}
\begin{figure}[bt]
\centering
\includegraphics[width=0.8\textwidth]{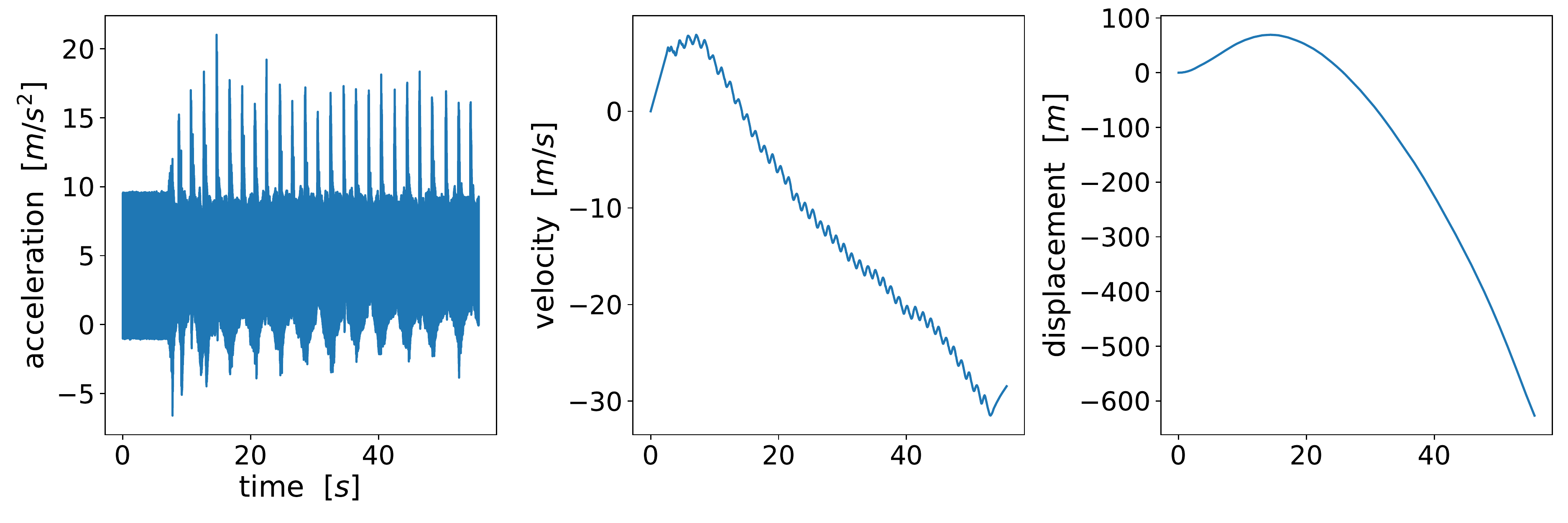}
\caption{Left: raw acceleration data, Middle: reconstructed velocity profile, Right: reconstructed acceleration profile. There is a large drift for velocity and displacement as no periodic patterns can be observed. A reasonable reconstruction should at least show periodic patterns as acceleration profile.}
\label{fig:accel_vel_displacement}
\end{figure}

We retrieved the raw magnometer, gyroscope and accelerometer readings from the inertial measurement units (IMUs) attached to both wrists and hands. There are three options where we could fit ProMPs, either on displacement profile $s(t)$ or velocity profile $v(t)$ or acceleration profile $a(t)$. We also derived the velocity and displacement profiles using Attitude Heading Reference System Filter (Ahrs Filter) \cite{sensor_drift} as shown in Figure \ref{fig:accel_vel_displacement}. However, it could be observed that the reconstructed $v(t)$ and $s(t)$ were largely distorted due to the (double) integration over time. The distortion of the reconstructed signal arose from the inaccurate recordings from in orientation or acceleration, where the error was accumulated and amplified after performing double integration over time \cite{pose_est}. 
\\
Fortunately, the acceleration profile does not require any time integration, hence bears the highest accuracy among these three options. Besides, it well explains the motion difference among different trajectories. The motion difference could still be shown in the acceleration profiles during the motion phase. For these reasons, we directly fit ProMPs on raw acceleration data. One pitfall of directly using raw acceleration profiles is that the raw acceleration can vary if the initial poses are different. To avoid that we precisely instructed the subjects on how to place their hands at the start of each experiment.

\subsection{What is the reconstruction error using ProMPs?}\label{se:fit_result}
To show the quality of the fitted model of ProMPs on the dataset, we computed the $D_{KLS}$ between the reconstructed trajectories from ProMPs and the post-processed ones on different hyper-parameter settings. In Table \ref{tab:recon_loss}, we list the reconstruction error defined as:
\begin{equation}
\label{eq:reconstruction_loss}
L_{rec} = \frac{1}{ \left | o \right |  \cdot \left | p \right | \cdot \left | q \right | } \sum_{o,p,q} D_{KLS}(\bm{\tau_{o,p,q}} \| \bm{\tau_{o,p,q}^{\prime}}),
\end{equation}
where $t$ refers to the number of discrete time points after time normalization, i.e. phase $z$, and $\bm{\tau}$ and $\bm{\tau^{\prime}}$ are respectively the set of post-processed trajectories and the set of reconstructed trajectories fitted by ProMPs as shown in Section \ref{se:trajectory_prediction}. Additional hyper-parameters of ProMPs are the centers $c_{i}$ and the bandwidth $h_{i}$ in Equation \ref{eq:bf} respectively. In our case, we set $h_{i}$ as a function of the number of basis functions $M$, so that the tunable hyper-parameter is merely $i$. The setting of bandwidth of each basis function goes as $h = 0.2 \cdot (c_{i+1}-c_{i})^{2}$ and the centers $c$ are uniformly distributed between the phase of $[0,1]$. It can be seen that the reconstruction loss decreases with the increasing number of basis functions. Although even better reconstruction loss can be achieved by increasing $M$, it risks overfitting from a machine learning perspective. We discuss this point in Section \ref{se:disucssion}. From a machine learning perspective, the raw data is usually projected into a feature space of smaller dimension than raw data. 

\begin{table}[bth]
\centering
\begin{tabular}{c c}
\toprule
Number of basis functions $M$ & Reconstruction loss $L_{rec}$\\
\midrule
$5$ &  $0.207 \pm 0.049$ \\
$10$ &  $0.073 \pm 0.019$ \\
$15$ &  $0.034 \pm 0.011$ \\
$20$ &  $0.016 \pm 0.005$ \\
\bottomrule
\end{tabular}
\caption{Reconstruction loss with respect to number of basis functions in ProMPs. The reconstruction loss is computed using $D_{KLS}(\bm{\tau}||\bm{\tau^{\prime}})$. Here we show the mean and standard deviation on the reconstruction loss from three randomly chosen participants.}
\label{tab:recon_loss}
\end{table}

In the remaining experiments, we set the hyper-parameters of ProMPs as: $n=20, h=0.2 \cdot (20-1)^{-2}$, and the regularizer term $\lambda = 1e$-$6$.
\begin{figure}[t]
\centering
\includegraphics[width=0.8\textwidth]{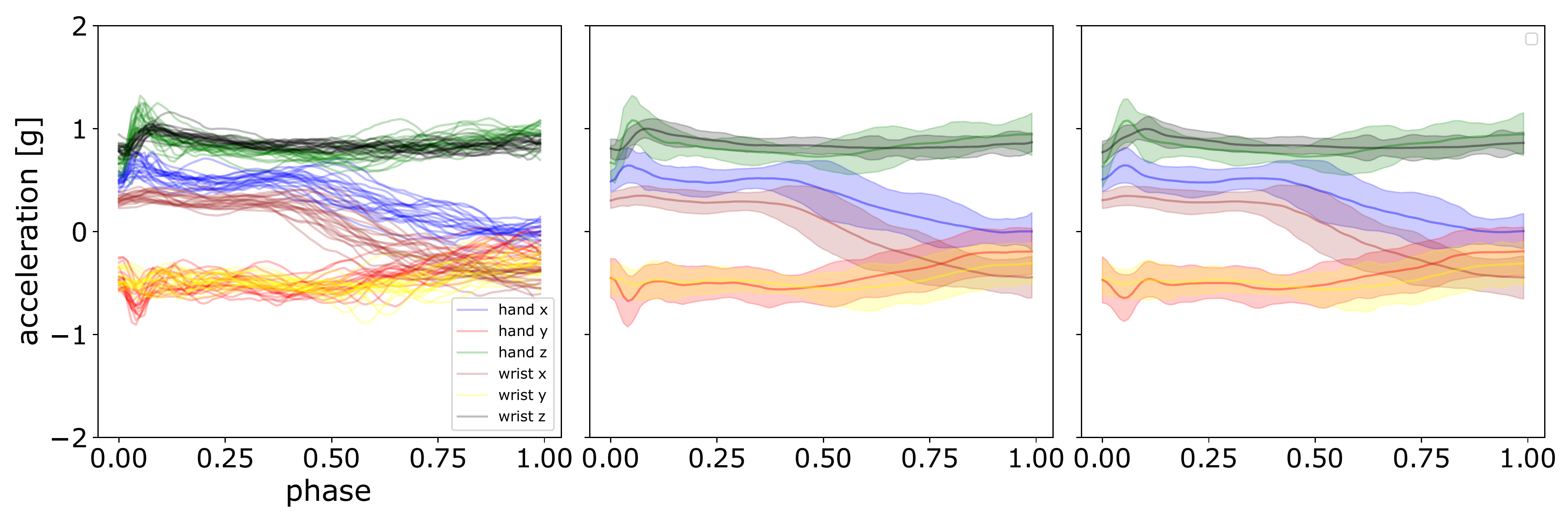}
\caption{Comparison of the mean and standard deviation of the post-processed acceleration profiles and the reconstructed acceleration profiles using ProMPs. The left panel shows the data after post-processing, the middle panel displays the mean and standard deviation of that data, and in the right panel, the reconstruction using ProMPs is illustrated. The shaded area corresponds to $2\sigma$ confidence interval.}
\label{fig:ProMP_illustration}
\end{figure}
An illustration using these parameters is shwon in Figure \ref{fig:ProMP_illustration}. It can be observed that ProMPs accurately model the recorded data.

\subsection{Can ProMPs be used to detect outliers?}\label{se:anomaly}
\begin{figure}[tb]
\includegraphics[width=0.98\textwidth]{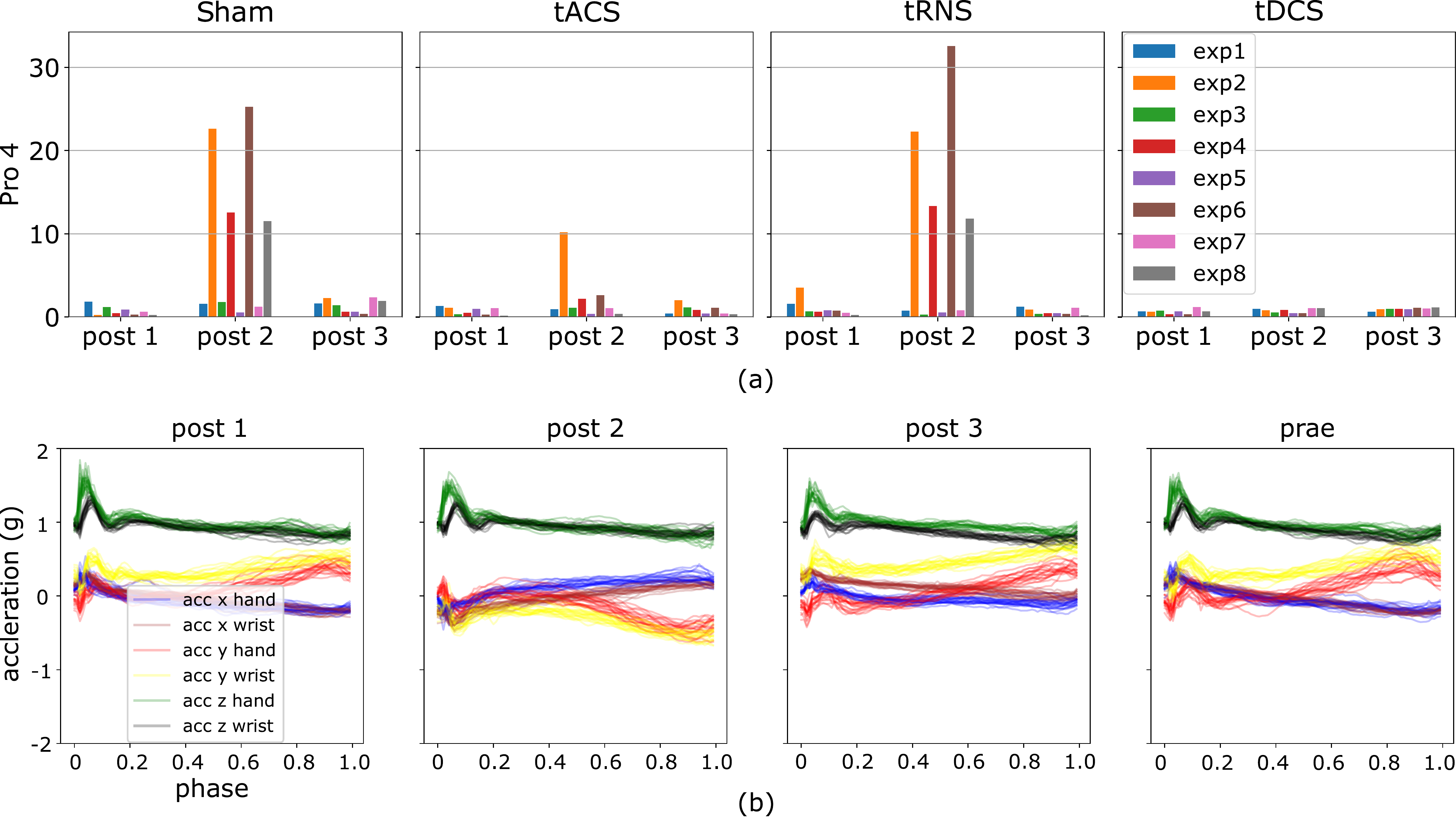}
\caption{Using ProMPs with symmetric KL-divergence helps detect potential data anomalies. (a) Potential anomalies in high KL-divergence case, i.e., high $D_{KLS}$ case (b) Flipping of the blue, red, brown and yellow curves, i.e., $x$, $y$-axis recordings in hand and wrist in \textit{post 2}. }
\label{fig:anomaly_pro004}
\end{figure}

We generated the $D_{KLS}$ in Equation \ref{eq:KLD_sym} for each participant and experiment according to the steps in Section \ref{se:kl_compute}. By analyzing the resulting statistics, we detected some potential anomalies. An initial observation is that $D_{KLS}$ is small among most of the experiments and subjects. However, there are peaked values in few cases, suspected as outliers. For instance, the irregular patterns (high divergent value) in the experiment $2$,$4$,$6$,$8$ in the day of Sham and tRNS for participant $4$ are present, as shown in Figure \ref{fig:anomaly_pro004}. A closer look into the post-processed data revealed that the $x$,$y$-axis recordings for both hand and wrist were flipped, a sign indicating the hand and wrist IMUs were misplaced. We also verified this hypothesis by looking into the recorded video and confirm the sensor misplacement. Similar problems of sensor misplacements were also observed in participant $11$ in experiment $2$,$4$,$6$,$8$ in $post\; 1$, tDCS.

Besides man-made mistakes, there are also other high divergent cases, e.g shown in Figure \ref{fig:exploration_pro019}. We analyse the potential cause in Section \ref{se:discuss_effect}. To automatically detect the data anomalies, we used the $3\sigma$ rule, which is equivalent to a confidence interval of $99.7\%$. Outliers are detected if $D_{KLS} > \mu_{D} + 3\sigma_{D}$ for the further statistical analysis. Here $\mu_{D}$ refers to the mean value of $D_{KLS}$ averaged over all non-corrupted experiments and participants, and $\sigma_{D}$ is the corresponding standard deviation. By this we show that the ProMPs with \textit{symmetric KL-divergence} can effectively detect anomalies, which is much more efficient than manual investigation of $1600$ sets of motions.

\subsection{How are the effects of tRNS, tDCS and tACS over the finger-tapping motion?} \label{se:initial result}
In this section, we present the results on the motion difference under different \textit{Transcranial Current Stimulation} methods. The difference is quantified by $D_{KLS}$ and ProMPs following the procedure in Section \ref{se:kl_compute} and Section \ref{se:anomaly}. Figure S1 in Supplementary Material shows the statistics of $D_{KLS}$ of each participant and experiment, excluding the outliers and corrupted data.

\begin{figure}[t]
\centering
\includegraphics[width=0.9\textwidth]{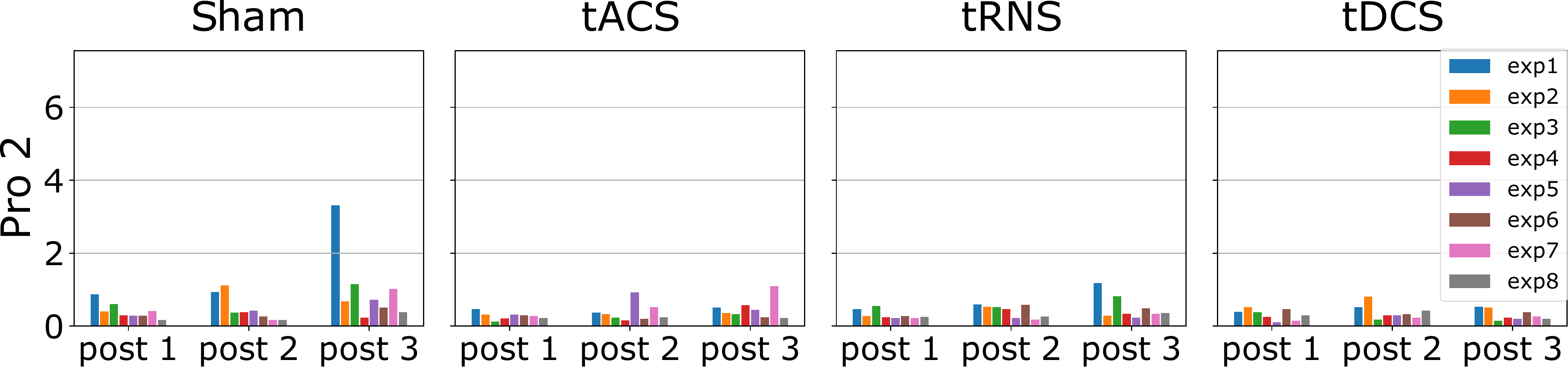}
\caption{This Figure shows the $D_{KLS}$ value of experiment subject $2$ for each stimulation approach, each experiment configuration and each post-stimulation phase. The $D_{KLS}$ are computed from the outward movements. The x-axis denotes the comparison between post-stimulation phase \textit{Post} and non-stimulated phase \textit{Prae}. The statistics already excludes the outliers via the $3\sigma$ rule mentioned in Section \ref{se:anomaly}. As the $D_{KLS}$ is quite low, no significant effect of tACS, tRNS and tDCS over the hand motion can be observed in this subject.}
\label{fig:KL_result_pro_002}
\end{figure}

We illustrate the statistics from one random participant in Figure \ref{fig:KL_result_pro_002}. By looking at the figure, the value of $D_{KLS}$ in most cases are around $1$. According to Figure \ref{fig:KL_value_correspondence}, the trajectory difference is minor in the case of $D_{KLS} \leqslant 2.5$. In this participant, the divergent value implies no pronounced discrepancy is observed in each type of stimuli or different post-stimulation phases. In addition, we check whether the effect of stimuli varies across the post-stimulation phases of $45$, $90$ and $135$ minutes after the stimulation is activated. In this participant, we did not observe any notable trend among different post-stimulation phases. Furthermore, the divergent values in Sham are similar to those in tACS, tDCS and tRNS for all $8$ experiments and post-stimulation phases. This indicates different stimulation approaches does not exert notable effects on movements on this experimental subject. The divergent value of $1$ mainly comes from the stochasticity of the participant's movement, i.e., the participant can not reproduce the exact same movement in different trials. 

We show a comprehensive result on all participants in Figure S1 in Supplementary Material, and further computed the mean and the standard deviation of the $D_{KLS}$ value in that figure over all participants and all $8$ experiment configuration based on both inward and outward movements respectively. The result is shown in Table \ref{tab:KL_div_averaged_summary_mvt12}. It can be firstly observed that the $D_{KLS}$ measure displays similar value between inward and outward movements when one compares each entry in two sub-tables. This result is anticipated, and indicates no significant difference can be captured for inward and outward movements. The inward and outward movements were segmented from one complete recordings as mentioned in Section \ref{se:data_process}.

Moreover, a persistent increasing trend in motion difference can be observed with the 
time after stimuli for each simulation method including the baseline approach Sham. This indicates the effect becomes more pronounced within the time interval of $135$ minutes after the stimuli is activated. However, the baseline approach Sham also exhibits similar tendency and divergent value as three other stimulation approaches. This implies that the hand trajectory in phase \textit{Prae} deviates to some extent from the hand trajectory in phase \textit{Post 1-3} not due to the stimuli. The standard deviation of the divergent values also increases with the time after stimulation for each stimuli type except Sham. We present an analysis on these results in Section \ref{se:disucssion}.

\begin{table}[tb]
\centering
\begin{tabular}{*7c}
\toprule
Type of stimuli &  \multicolumn{2}{c}{$D_{KLS}(\bm{\tau^{\prime}_{prae}} || \bm{\tau^{\prime}_{post\: 1}})$} & \multicolumn{2}{c}{$D_{KLS}(\bm{\tau^{\prime}_{prae}} || \bm{\tau^{\prime}_{post\: 2}})$}&  \multicolumn{2}{c}{$D_{KLS}(\bm{\tau^{\prime}_{prae}} || \bm{\tau^{\prime}_{post\: 3}})$}\\
{}   & mean   & std dev    & mean   & std dev   & mean   & std dev\\
\midrule
Sham   &  0.917 & 1.054   & 0.910  & 0.545   &  1.078  & 0.690  \\
tDCS   &  0.680 & 0.554   & 0.815  & 0.630  &  0.938 & 0.892\\
tACS   &  0.744  & 0.669   & 0.988  & 0.747   & 1.069  & 0.852\\
tRNS   &  0.802  & 0.719   & 0.947  & 0.806   & 1.064  & 1.072\\
\bottomrule
\end{tabular}

\bigskip

\centering
\begin{tabular}{*7c}
\toprule
Type of stimuli &  \multicolumn{2}{c}{$D_{KLS}(\bm{\tau^{\prime}_{prae}} || \bm{\tau^{\prime}_{post\: 1}})$} & \multicolumn{2}{c}{$D_{KLS}(\bm{\tau^{\prime}_{prae}} || \bm{\tau^{\prime}_{post\: 2}})$}&  \multicolumn{2}{c}{$D_{KLS}(\bm{\tau^{\prime}_{prae}} || \bm{\tau^{\prime}_{post\: 3}})$}\\
{}   & mean   & std dev    & mean   & std dev   & mean   & std dev\\
\midrule
Sham   &  0.874 & 0.871   & 0.900  & 0.624   &  1.122  & 0.903  \\
tDCS   &  0.741 & 0.695   & 0.803  & 0.676  &  0.925 & 0.912\\
tACS   &  0.749  & 0.771   & 0.980  & 0.800   & 1.044  & 0.860\\
tRNS   &  0.827  &  0.733   & 0.943  & 0.869   & 1.050  & 1.024\\
\bottomrule
\end{tabular}
\caption{A comparison of mean and standard deviation $D_{KLS}$ on both inward (upper table) and outward movements (lower table) over all the available statistics in Figure S1 in Supplementary Material. The outliers have been excluded. The mean and standard deviation are computed by collapsing the dimension of number of participants and $8$ experiment configurations.}
\label{tab:KL_div_averaged_summary_mvt12}
\end{table}

\subsection{Can time-specific differences be detected on a millisecond time scale?} \label{se:sliding window}
\begin{figure}[tb]
\centering
\includegraphics[width=0.98\textwidth]{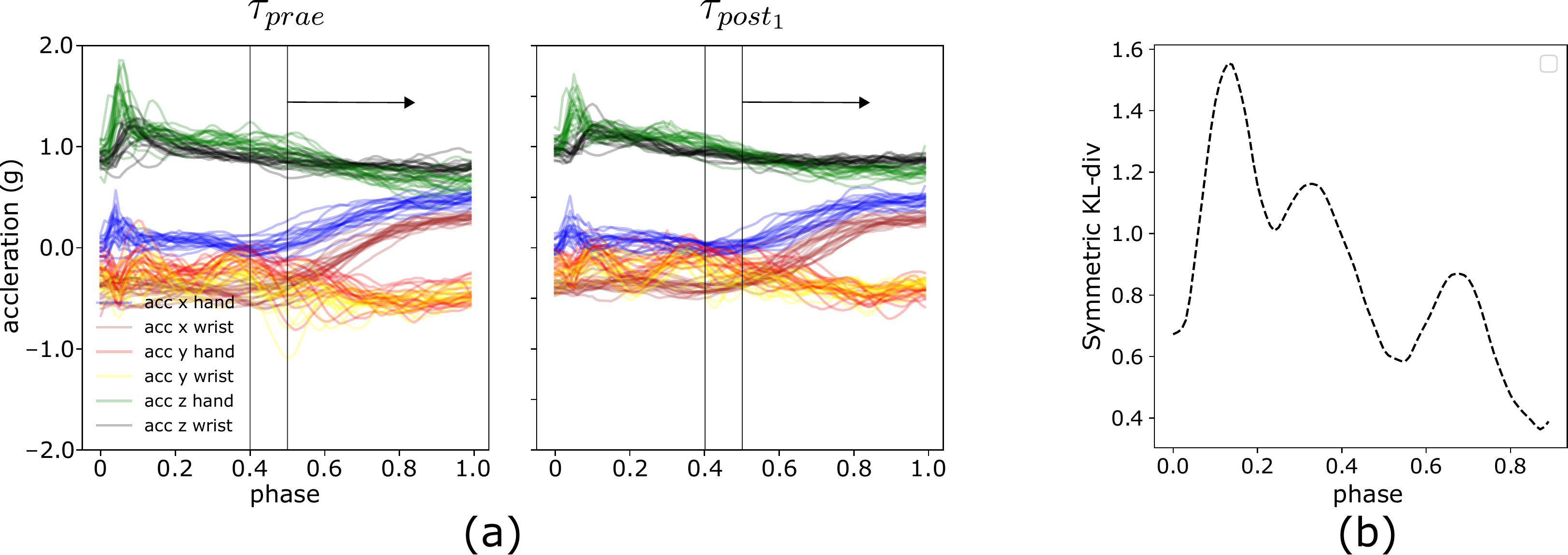}

\caption{An illustration of sliding window approach to reveal time-specific difference in trajectory space. (a) Sliding window with the size of $1/10$ of the total normalized time, i.e., \textit{phase} between $\bm{\tau}_{prae}$ and $\bm{\tau}_{post\;1}$ from one random experiment (b) The resultant symmetric KL-divergence averaged over all $6$ axes between two experiments in sliding window manner.}
\label{fig:sliding window}
\end{figure}

As a step further, one can also examine which part the trajectories contributes to the motion difference by using a sliding window approach. The window size is pre-defined and we compute the $D_{KLS}$ within the window, slided across the \textit{phase} $z_{t}$. The resultant curve illustrates the general trend of difference in relation of time. The sliding window approach can extract the most significant difference or features between two sets of trajectories as demonstrated in Figure \ref{fig:sliding window}. The most notable difference is observed within $10\%$ to $20\%$ of the \textit{movement phase} in Figure \ref{fig:sliding window}(b). This phase corresponds in our experiments to the hand lifting. We provide a comprehensive outlook on the trend of trajectory divergence using sliding window approach in Figure \ref{fig:sliding_window_mvt12} and Figure \ref{fig:sliding_window_mvt21} in Supplementary Material for each subject and each experiment.

\section{Discussion}\label{se:disucssion}
In this section, we first discuss the advantages and limitations of our approach ProMPs for modelling human motions and also shed light on the effects of the \textit{non-invasive brain stimulation} methods tACS, tRNS and tDCS on the finger-tapping motions.

\subsection{Efficiency of ProMPs in Motion Modelling} \label{se:fast_ProMP}
As an auxiliary analysis tool, ProMPs with \textit{symmetric KL-divergence} provides an efficient way of motion analyses compared to manual investigations. The computational time needed for learning ProMPs and computing the \textit{symmetric KL-divergence} takes less than $10$ minutes for all $1600$ experiments, i.e., $1600 \cdot 20 = 32000$ strokes of movements. The duration of $10$ minutes also includes the time spent for the expensive frequent read write processes, i.e., loading post-processed data and saving the statistics. In contrast, the entire analysis time could endure days for the same amount of data in human case. 

\subsection{ProMPs with Symmetric KL Divergence as a Consistent and Robust Distance Metric} \label{se:consistent_ProMP}
We also show that ProMPs together with \textit{symmetric KL-divergence} constitute a consistent metric in quantifying the difference between sets of trajectories. We compared the $D_{KLS}$ value respectively for inward and outward movements both on each experiment as shown in Figure S1 in Supplementary Material. We further computed the mean and standard deviation by collapsing all the dimensions, i.e., experiment configurations, the number of experimental subjects, stimulation approaches and post-stimulation phases. The result is shown in Table \ref{tab:temps}. As a consistent and robust distance metric, they should hold similar outcomes.

As can be seen from Figure S1 in Supplementary Material and Table \ref{tab:temps}, the \textit{symmetric KL-divergence} is highly similar for all inward and outward movements. The similarity is expected as the divergent values $D_{KLS}$ measured from inward and outward movements should be close to each other. This also verifies the consistency of ProMPs and $D_{KLS}$ as a measure of difference between trajectories. 

\begin{table}[h]
\centering
\begin{tabular}{c  c}
\toprule
&Averaged $D_{KLS}$ \\
\midrule
Inward movements & $0.874 \pm 0.689$ \\
Outward movements & $0.865 \pm 0.695$ \\
\bottomrule
\end{tabular}
\caption{A comparison of the averaged \textit{symmetric KL-divergence} on both inward and outward movements over all experiments and participants, the mean and standard deviation are computed by averaging each $D_{KLS}$ values in Figure S1 in Supplementary Material respectively.} 
\label{tab:temps}
\end{table}

Another benefit is the robustness of ProMPs against noise as a feature extractor, which is critical for further analysis of medical methods. In some manually-designed features, the noise extensively affects the feature extraction and consequentially leads to wrong classification or distance metric computation. For instance, in \cite{bologna2016effects}, they define the maximal velocities and others from the complete trajectory as the features, which are highly-sensitive to noise.

\begin{figure}[tb]
\includegraphics[width=0.99\textwidth]{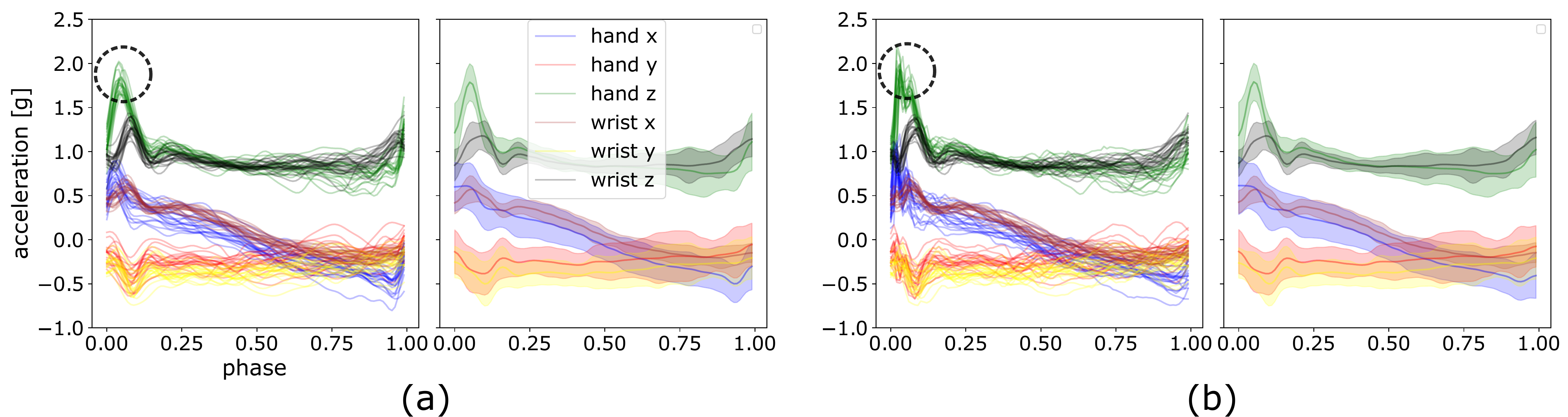}
\caption{We provide a set of post-processed trajectories with filtering (a) and without filtering (b). Frequencies higher than $20\; hz$ are filtered out. A high similarity between fitted results using ProMPs can be observed, especially the automatic filtering effect can be seen from the green curve at its peak value, marked in circle. }
\label{fig:filter_comp}
\end{figure}

In ProMPs, however, the noise is filtered through applying the basis functions. The robustness against noise stems from the shape of basis functions and the number of the basis functions used. High-frequency noise can be interpreted as a non-smooth jerk, which cannot be perfectly fitted using a limited number smooth radial basis functions. Therefore, the noise is automatically left out in the process of model learning with a suitable choice on model hyper-parameter. Note that by having a extremely small bandwidth of the basis functions (see $h$ in Equation \ref{eq:bf}) and a large number of $M$, it is theoretically possible to even fit to noise, i.e., the model can overfit. In ProMPs, high frequency artefacts are typically not pronounced. In contrast, additional step of filtering is always necessary in frequency-domain models. Figure \ref{fig:filter_comp} illustrates a visual comparison of non-filtered data and a filtered one, where one can observe a noise-filtering effect on the reconstructed trajectories. Moreover, the reconstructed trajectories using filtered data is highly similar to the ones using non-filtered data, which is desired. Given sufficient demonstrations, ProMPs is also robust against outliers, as the mean and standard deviation will not be shifted greatly for a set of demonstrations.

\subsection{Generalizability of ProMPs to other Motion-Modelling Tasks} \label{se:generalizability_ProMP}
One clear advantage using ProMPs compared to other manual feature design approaches is its generalizability to out-of-domain tasks only with minimal efforts in hyper-parameter tuning. 
The center $c_{i}$ and the bandwidth $h_{i}$ of each basis function are automatically adapted according to the number of basis functions $M$. The only true open parameter is $M$, which is adapted based on the non-linearities or complexities in the data. Typical values for $M$ are in the range of $10$ to $30$. The key of ProMPs' generalizability to other task lies in the extraction of key features representing the trajectory shape, which allows to quantify similarities or to visualize clusters. In \cite{rueckert2015extracting}, the model was extended to also learn latent robot control parameters in a two-dimensional space. Such approaches can be also used for visualizing and analyzing complex human motion data.

\subsection{Limitations using ProMPs} \label{se:disadv_ProMP}
In this part, we also discuss the limitations of ProMPs. First, a proper selection of hyper-parameter is necessary to achieve desired feature extraction. This is also referred to as the general challenge in machine learning, the mean-variance tradeoff. As can been seen in Table \ref{tab:recon_loss}, an insufficient number of basis function $M$ can cause a high bias, whereas high model complexity risks modelling undesired motion features, i.e., noise in extreme case. Another point is that at least two trajectories must be present for fitting ProMPs to enable mean-variance scheme from a theoretical perspective. In theory, it is always better to have as plentiful demonstrations as possible. From a statistical view, more samples drawn from an unknown probability distribution leads to better approximation of the ground-truth distribution. This in turn poses the requirement that the tasks should be performed repetitively. However, in practice typically $10$ to $30$ repetitions are sufficient. 

It is also noteworthy that ProMPs require the time-alignment or normalization to make each segmented data have the same length. By performing time alignment and normalization, the model diminishes the difference between trajectories by ignoring the actual duration of each segments. 

\subsection{The Effects of tACS, tDCS and tRNS on Finger-Tapping Motions} \label{se:discuss_effect}
In this work, the effects of tACS, tDCS and tRNS are quantified using ProMPs and $D_{KLS}$. If the divergent values of the stimulation approaches are different from the baseline approach Sham, the effect of the stimuli on the motion can be indeed inferred. However, the reults shown in Section \ref{se:initial result} shows no specific effects can be seen from tACS, tDCS or tRNS using our approach.

\begin{figure}[tb]
\centering
\includegraphics[width=0.99\textwidth]{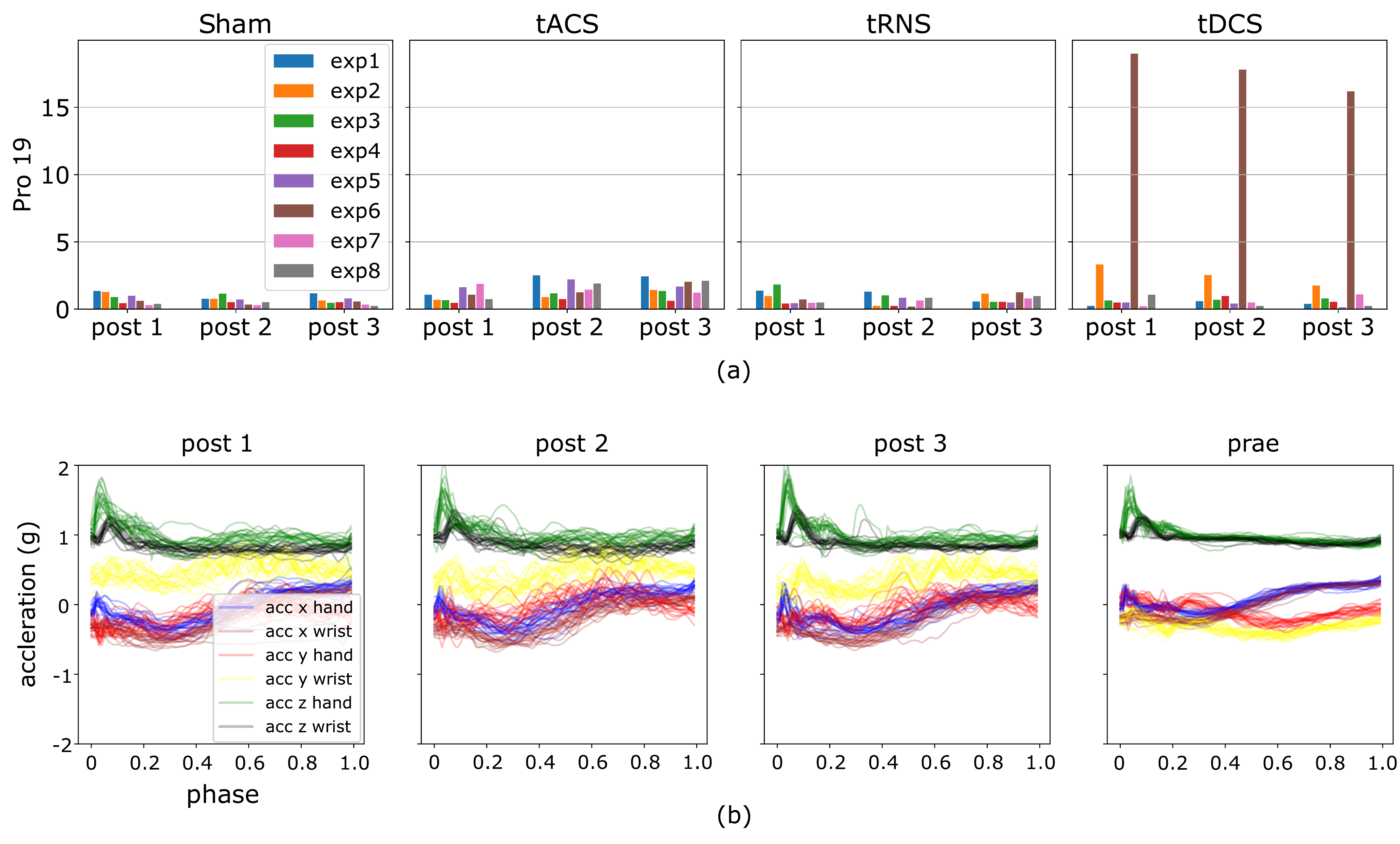}
\caption{An illustration of ProMPs with $D_{KLS}$ detecting data outliers. (a) High $D_{KLS}$ case in participant $19$, tDCS, the sixth experiment, brown bar shows remarkably higher value than others, indicating a potential outlier. (b) Post-processed data for the experiment $6$ in tDCS. The yellow curve accounts dominantly for the high divergence, interpreted as a different motion behavior in \textit{Prae} from other three.}
\label{fig:exploration_pro019}
\end{figure}

Nevertheless, before applying the $3\sigma$ rule to exclude outliers, we indeed saw a few individual cases of high $D_{KLS}$. For instance, we observed a peaked $D_{KLS}$ value in the sixth experiment in tDCS from participant $19$ as shown in Figure \ref{fig:exploration_pro019}. It shows that the post-processed trajectory in \textit{Prae} displays a different pattern from \textit{Post 1-3}, whereas the trajectories in different post-stimulation phases are similar. Nonetheless, the difference is only noted in the sixth experiment in this participant, whereas the divergent values for all other experiments are similar to each other and well below the one in the sixth experiment. We deemed such case as outliers, and it was automatically ruled out with $3\sigma$ criterion.

The potential cause of such high divergent values is that a participant attempted to finish the experiment in different manner or the participant's initial pose for each experiment is deviating, despite the fact that each participant was told to start with the same initial pose and same motion behavior for each experiment as much as possible. 
Some subjects might still be in the process of getting familiar with the experiments so that they performed exploratorily between experiments. In some cases, they forgot the initial pose after the interval of around $45$ minutes between each post-stimulation approach so that the acceleration profiles in local coordinate frame exhibit different patterns. Since the whole procedure of recording and accepting stimulation and conducting the experiment endures a complete afternoon, the participants might experience tiredness, attention deficiency or other factors, leading to the undesired points mentioned above. We regarded such kind of peaked $D_{KLS}$ cases as outliers and did not include them in the final analysis.

Moreover, we noticed an persistent but very slight increase in divergent value for all stimuli types including Sham with the increasing post-stimulation periods. The overall slight increasing trend of $D_{KLS}$ both in mean and standard deviation can be interpreted as the tiredness and loss of attention from the experimental subjects after the long experiment duration, i.e., more than $4$ hours. 

However, we do not conclude that tACS, tDCS or tRNS plays no an effect on human motions in this work for several reasons. Firstly, the data collection is still on-going, it would be statistically more sensible to have more data. We planned to integrate $30$ participants at the end. Another point is that ProMPs cannot capture the effect of the time duration of each motion segments, on which these brain stimulation approaches could take effect. According to some other work \cite{orru2020clinical, benussi2017long}, tCS approaches could also influence the accuracy of tapping, i.e., whether the finger taps the middle point of the touching pad. The subjects could potentially adjust the trajectory with the presence of the plasticity effect, where the fitted acceleration profile can only partially show this adjustment effect. The ideal case to examine this is the montage of additional sensors to detect the touching point or an accurate reconstruction of the displacement profile in absolute coordinate system.

\section{Conclusions}\label{se:Conclusions}
In this paper, we proposed \textit{Probabilistic Movement Primitives} (ProMPs), a well-established approach in robot skill learning, for human motion analysis to examine how variants of \textit{Transcranial Current Stimulation} approaches affect human motions in finger-tapping experiments based on IMU recordings. We showed ProMPs together with the \textit{symmetric Kullback-Leibler divergence} constituted a robust distance metric for measuring the difference between sets of motion trajectories. As an auxiliary analysis tool, ProMPs first provide much faster diagnosis and more objectivity than human experts and can even reveal potential experimental mistakes, e.g., sensor misplacement. Compared with other methods that rely on manually-designed features or frequency-domain analysis, ProMPs are robust against noise and extract features of the trajectory shape, which makes it easily extendable to other tasks. In ProMPs, merely a single model hyper-parameter, namely the number of basis functions, needs to be defined or determined. In this paper, we discussed how this parameter can be determined by analyzing the reconstruction error. We also demonstrated and discussed how ProMPs can be used for filtering noise which tremendously reduced the required human effort for data post-processing. 
 
In current work, we discussed the effect of transcranial random noise stimulation (tRNS), transcranial alternating current stimulation (tACS) and transcranial direct current stimulation (tDCS) on finger-tapping motions using an initial data of $10$ participants. In our initial study, our approach did not reveal any significant effects of these stimulation approaches on the tapping movement. Additional evaluations with more participants are needed for in-depth investigations, which is part of future work.

\section*{Acknowledgments}
The project receives funding from the Deutsche Forschungsgemeinschaft (DFG, German Research Foundation) – No 430054590 (TRAIN, to ER) and (DFG, WE 5919/2-1 to AW) and the Else Kröner-Fresenius Foundation (2018\_A55) to AW. The authors also show great appreciation to Julius Verrel for the code for data post-processing parts, and also Nils Rottmann and Ralf Bruder for their suggestions on the data post processing. 

\bibliographystyle{unsrt}  
\bibliography{references}

\pagebreak
\begin{center}
\textbf{\large Supplemental Materials}
\end{center}
\setcounter{equation}{0}
\setcounter{figure}{0}
\setcounter{table}{0}
\setcounter{page}{1}
\makeatletter
\renewcommand{\theequation}{S\arabic{equation}}
\renewcommand{\thefigure}{S\arabic{figure}}

In this section, we elaborated the details of the experimental schedule within each day and the randomized order of each participant on accepting the stimulation approaches of tACS, tDCS, tRNS and Sham. We also showed how each unit experiment was configured. In addition, a complete statistics on characterizing the effects of stimulation using $D_{KLS}$ with ProMPs mentioned in Section 3, and the time-specific difference with a sliding window approach on each participant, experiment and post-stimulation phase was illustrated.

\begin{table}[htpb]
\centering 
\begin{tabular}{c c c c c}
\toprule
 \makecell{Time w.r.t.\\stimuli} & \makecell{Day 1\\ Sham} & \makecell{Day 2\\ tACS} & \makecell{Day 3\\ tRNS} & \makecell{Day 4\\ tDCS}\\
\midrule
-$90$ min &Erst&Erst&Erst&Erst\\
-$45$ min &  Prae & Prae & Prae & Prae\\
$35$ min &  Post 1 & Post 1 & Post 1 & Post 1\\
$75$ min &  Post 2 & Post 2 & Post 2 & Post 2\\
$115$ min &  Post 3 & Post 3 & Post 3 & Post 3\\
\bottomrule
\end{tabular}
\caption{Illustration of experimental schedule of participant $1$, negative value means prior to an event, positive value means post to an event. Note each day, i.e., Day 1, Day 2, Day 3 and Day 4 are separated from each other by one week. We also instructed every participant to stick to the same motion pattern as much as possible to avoid the effect of exploring different motion patterns to finish the experiment, i.e., reduce the trajectory variance not caused by stimuli.}\label{tab:exp_schedule}
\end{table}

\begin{table}[htpb]
\begin{center}

\begin{tabular}{c c c c}
\toprule
 \makecell{Experiment \\ index} & \makecell{Tapping direction} & \makecell{Rhythmic / Rapid \\ motion} & \makecell{Right / Left \\ hand}\\
\midrule
$1$ &  left-right & rhythmic & right hand\\
$2$ &  left-right & rhythmic & left hand\\
$3$ &  left-right & rapid & right hand\\
$4$ &  left-right & rapid & left hand\\
$5$ &  forward-backward & rhythmic & right hand\\
$6$ &  forward-backward & rhythmic & left hand\\
$7$ &  forward-backward & rapid & right hand\\
$8$ &  forward-backward & rapid & left hand\\
\bottomrule
\end{tabular}
\caption{A detailed illustration on the configuration of eight finger-tapping experiments.}\label{tab:8_exp}
\end{center}
\end{table}

\begin{table}[bthp]
\centering
\begin{tabular}{c c c c c}
\toprule
 \makecell{Subject index} & Day 1 & Day 2 & Day 3 & Day 4\\
\midrule
1 & Sham & tACS	& tRNS & tDCS\\
2 &  tACS & Sham & tRNS & tDCS\\
4 &  tDCS&	Sham&	tRNS&	tACS\\
7 &  tRNS&	tACS&	tDCS&	Sham\\
8 &  tDCS&	Sham&	tACS&	tRNS\\
9 &  Sham&	tDCS&	tACS&	tRNS\\
11 & tACS&	tRNS&	tDCS&	Sham\\
13 & tDCS&	tACS&	Sham&	tRNS\\
15 & tACS&	tRNS&	Sham&	tDCS\\
19 & tACS&	tDCS&	Sham&	tRNS\\
\bottomrule
\end{tabular}
\caption{Shuffled order on stimulation approach. Note there are skips between subject indices as the experiments of the subject with the skipped indices are still on-going.}
\label{tab:exp_schedule2}
\end{table}

\begin{figure}[bhtp]
\begin{center}
\includegraphics[width=\textwidth,height=0.9\textheight,keepaspectratio]{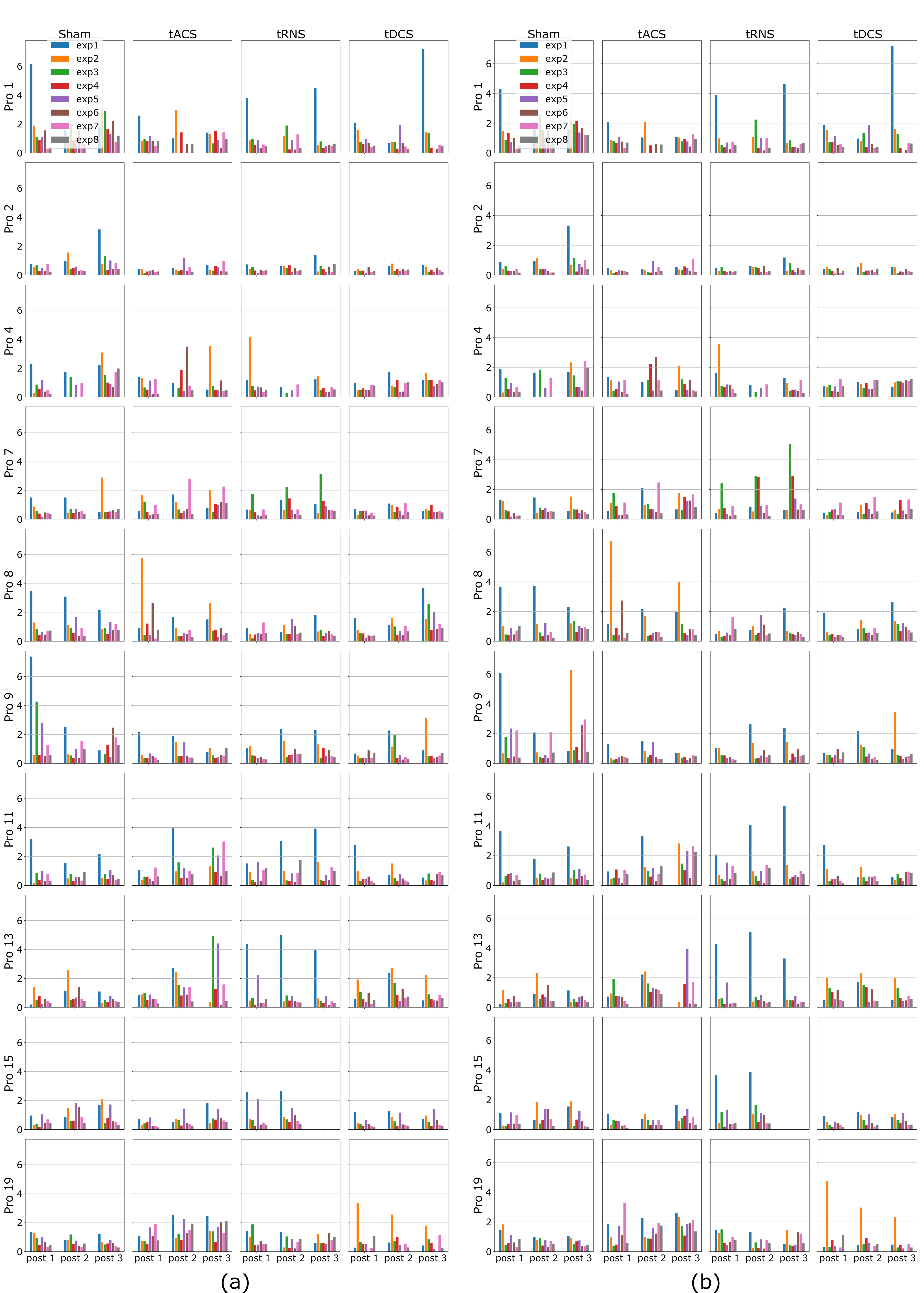}
\caption{Trajectory difference measured by ProMPs and \textit{symmetric KL-divergence} on $10$ participants. The result is demonstrated in terms of each experiment setting, stimulation method and each phase after stimulation. (a) Distance measured on all the segmented inward movements for arm (b) Distance measured on all the segmented outward movements for arm. The missing bars correspond to the filtered outliers.}\label{fig:initial_result}
\end{center}

\end{figure}

\begin{figure}[h]
\begin{center}
\includegraphics[width=\textwidth,height=0.9\textheight,keepaspectratio]{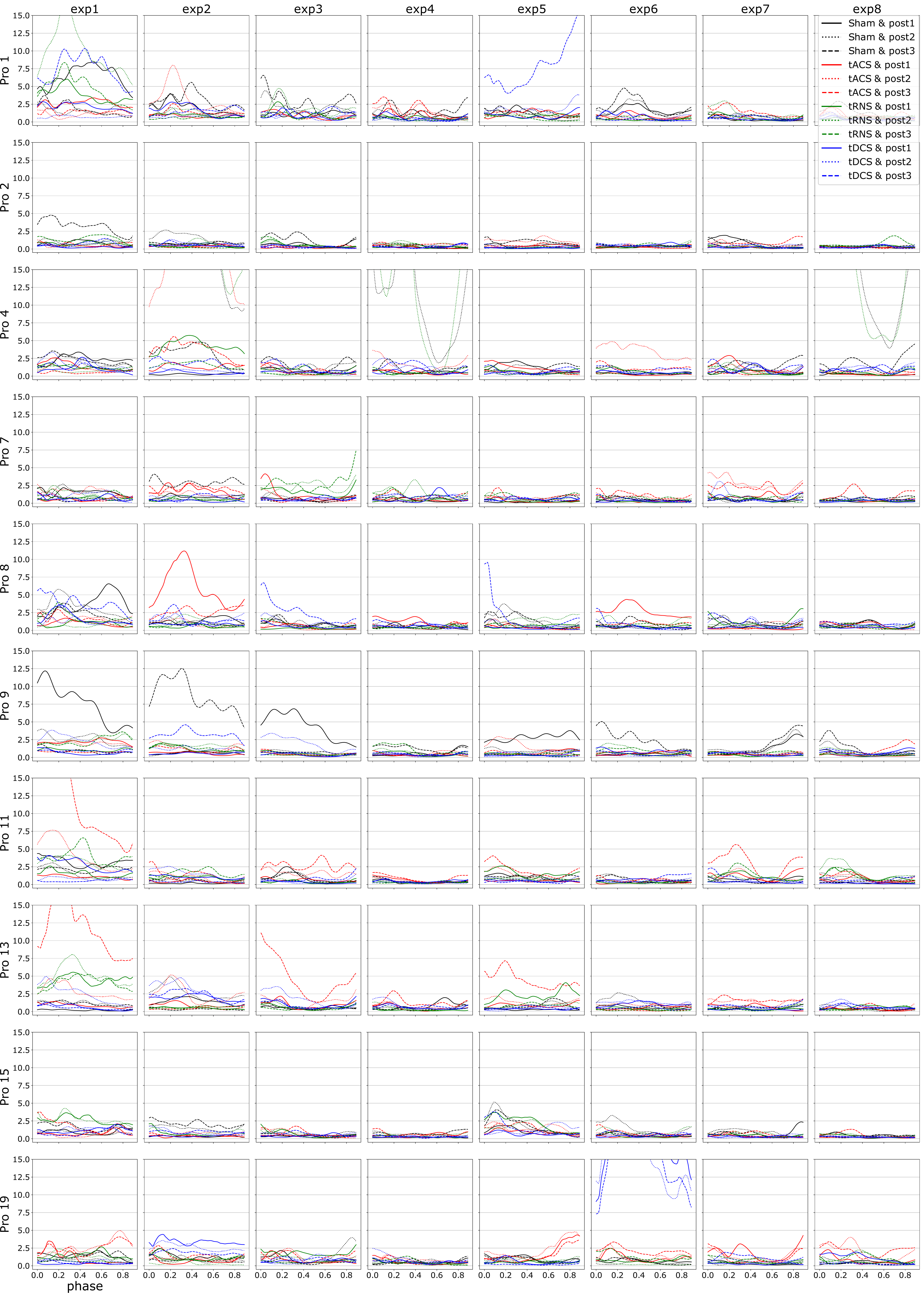}
\caption{Sliding window for time-specific trajectory difference on all inward arm  movements, measured via ProMPs and \textit{symmetric Kl-divergence}.}\label{fig:sliding_window_mvt12}
\end{center}
\end{figure}

\begin{figure}[h]
\centering
\includegraphics[width=\textwidth,height=0.9\textheight,keepaspectratio]{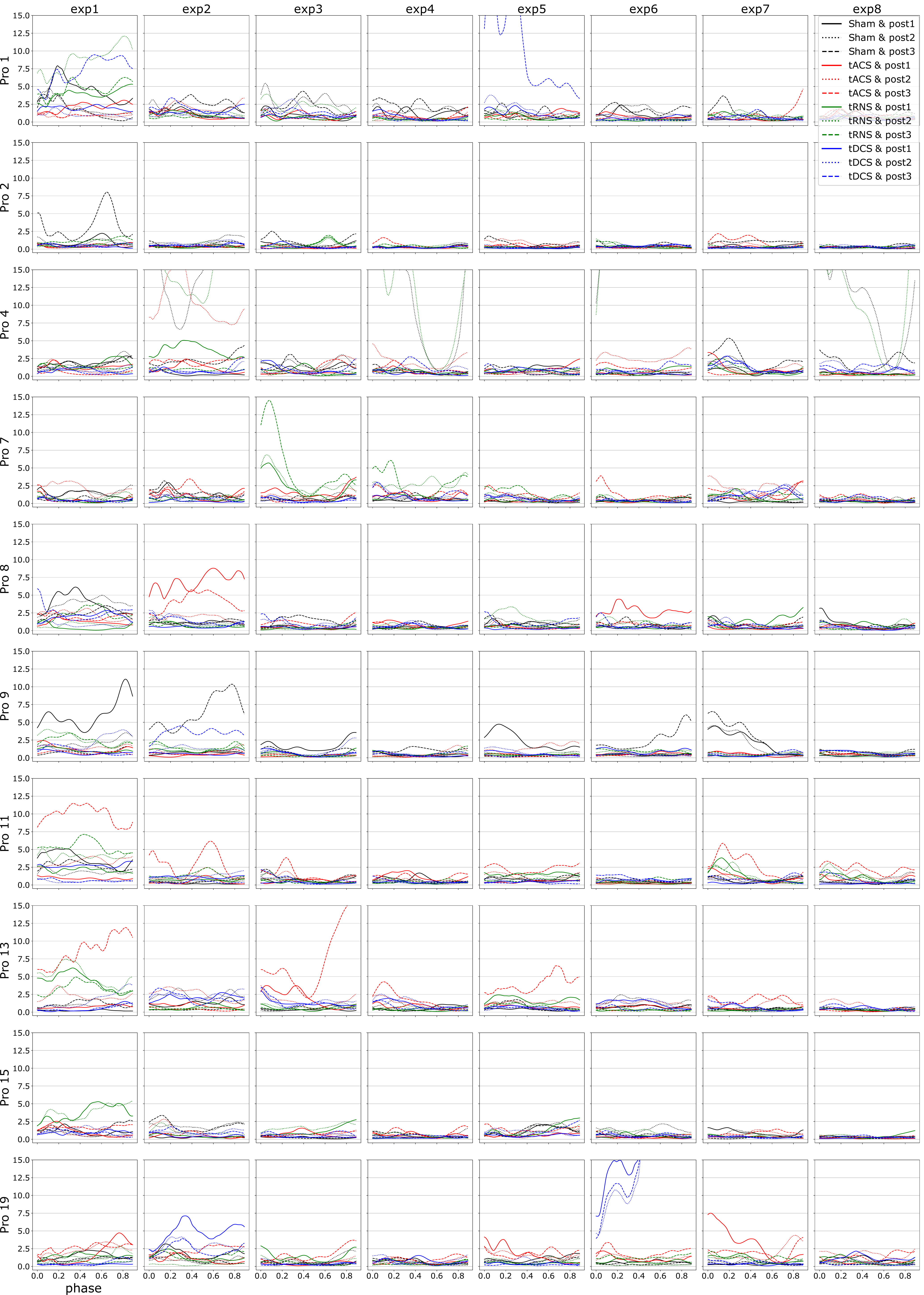}
\caption{Sliding window for time-specific trajectory difference on all outward arm  movements, measured via ProMPs and \textit{symmetric Kl-divergence.}}
\label{fig:sliding_window_mvt21}
\end{figure}

\end{document}